\documentclass{article}

    \PassOptionsToPackage{numbers, compress}{natbib}

    \usepackage[preprint]{neurips_2019}

\usepackage[utf8]{inputenc} 
\usepackage[T1]{fontenc}    
\usepackage{hyperref}       
\usepackage{url}            
\usepackage{booktabs}       
\usepackage{amsfonts}       
\usepackage{nicefrac}       
\usepackage{microtype}      
\usepackage{graphicx}
\usepackage{subcaption}
\usepackage{amsmath}
\usepackage{amssymb}
\usepackage{tabulary}
\usepackage{dsfont}
\usepackage{tabularx}

\title{DoorGym: A Scalable Door Opening  Environment and Baseline Agent}

\author{Yusuke~Urakami$^1$,  Alec~Hodgkinson$^1$, Casey~Carlin$^1$, Randall~Leu$^1$, Luca~Rigazio$^{1,2}$\\
       $^1$Panasonic Beta, CA, USA $^2$Totemic Inc., CA, USA\\
       \texttt{yusuke.urakami@us.panasonic.com}\\ 
       \And
        Pieter~Abbeel\\ University of California Berkeley, USA\\
        \texttt{pabbeel@cs.berkeley.edu}
       }

\begin{document}

\maketitle

\begin{abstract}
In order to practically implement the door opening task, a policy ought to be robust to a wide distribution of door types and environment settings. Reinforcement Learning (RL) with Domain Randomization (DR) is a promising techniques to enforce policy generalization, however, there are only a few accessible training environments that are inherently designed to train agents in domain randomized environments. We introduce DoorGym, an open-source door opening simulation framework designed to utilize domain randomization to train a stable policy. We intend for our environment to lie at the intersection of domain transfer, practical tasks, and realism. We also provide baseline Proximal Policy Optimization and Soft Actor-Critic implementations, which achieves success rates between  0\% up to 95\% for opening various type of doors in this environment. Moreover, the real world transfer experiment shows the trained policy is able to work in the real world. Environment kit available here: https://github.com/PSVL/DoorGym/
\end{abstract}

\section{Introduction}

\begin{figure}[t]
    \centering
    \includegraphics[width=1.0\textwidth]{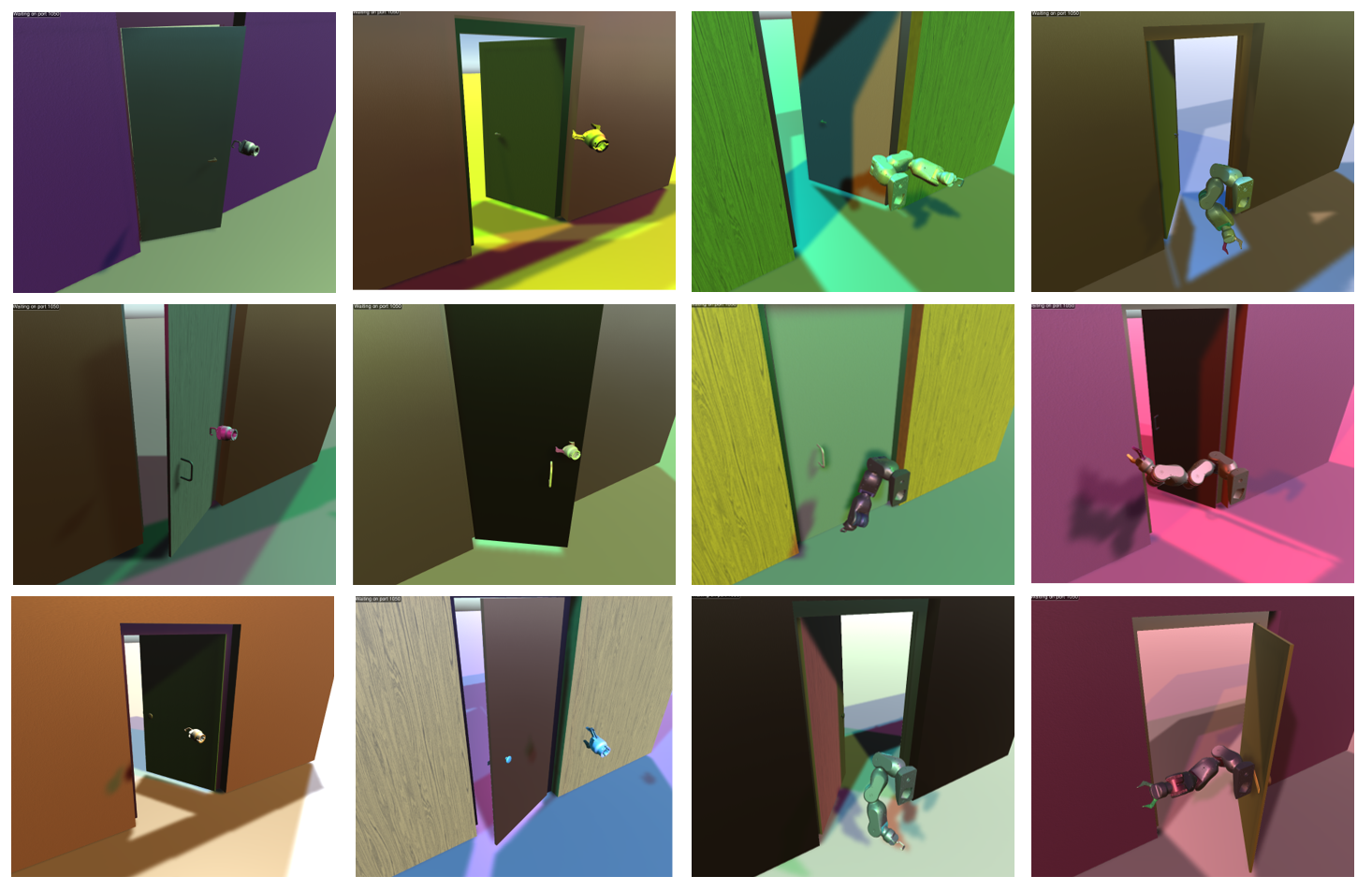}
    \caption{Various doors sampled with different visual and physical characteristics. For a full list of ramdomizable parameters, see table \ref{tab:randomized_doorworld_parameters}
    in the appendix.}
    \label{randomized_environment_figure}
\end{figure}



Door opening is a fundamental skill for any robot that interacts with humans. In the past, there have been a variety of approaches to solving the door opening task~\citep{5649847, du2015whatha, 8023522, 7989385, 6385835, endres2013, 1308857, 5174717, 6095096}. Prior work can be roughly categorized into a few different approaches: Door Keypoint Detection, Model Based, and RL based. Keypoint Detection approaches often involve locating the doorknob location, and axes of rotation, then using motion planning to open the door~\citep{5649847, 1308857, 5174717}. There have been several model based approaches that range from creating a model of the door~\citep{6385835} (Karayiannidis et al, 2012) to learning the kinematics of doors~\citep{endres2013}(Endres et al, 2013). Reinforcement Learning based approaches are beginning to gain more attention. Kalakrishnan et al, 2011~\citep{6095096} use compliant control and  $\textrm{PI}^2$~\citep{Theodorou:2010:GPI:1756006.1953033}(Theodorou et al, 2010) to open a door in a very narrow environment. Nemec et al, 2017~\citep{8023522} expanded the use of compliant control and $\textrm{PI}^2$ to transfer from simulator to multiple different doors, including a real door. More recently there have been some purely RL based approaches, though they have not been focused on opening doors. Both Gu et al, 2017~\citep{7989385} and Rajeswaran et al, 2017~\citep{DBLP:journals/corr/abs-1709-10087} have door opening environments. These research achieved an door opening task at certain level, however, they were not focusing on a generalization of a policy. A practical door opening policy must be robust to many different doors, lighting conditions, and different environment settings. 

\begin{figure}[t]
    \centering
    \includegraphics[width=1.0\textwidth]{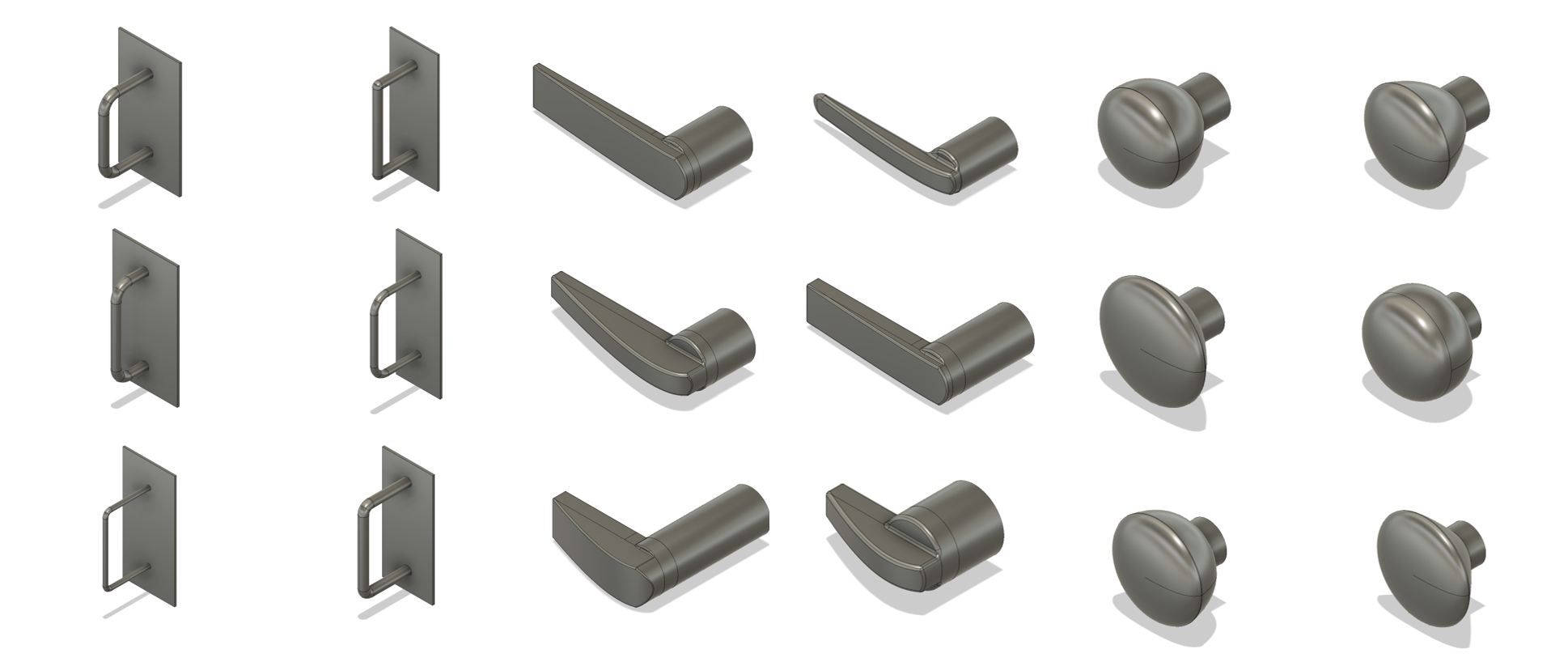}
    \caption{Random Samples from the doorknob distribution. 3000 each for knob are included in the door knob dataset.}
    \label{doorknob_figure}
\end{figure}

There are various techniques to make a policy robust to many different environments. Domain randomization (DR) is one such technique. DR assumes that it is hard to perfectly model the target domain, but it is easy to create many different simulations that approximate the target domain. With this assumption, it is possible to ensemble a variety of simulator environments with different visual or physical properties to generalize to a domain that overlaps the target domain. Rajeswaran et al, 2016~\cite{DBLP:journals/corr/RajeswaranGLR16} explored making policies more robust in the simple case of the MuJoCo Hopper and Half Cheetah environments using random physical properties. Sadeghi et al, 2017\citep{CAD2RLSadeghi} further explored using domain randomization with realistic textures to bridge the reality gap between simulation and the real world. This was done by retexturing CAD models to perform zero shot transfer for drone crash avoidance. Tobin et al, 2017~\citep{DBLP:journals/corr/TobinFRSZA17} abandons using realistic textures and instead uses synthetic textures to accomplish fine-grained tasks such as object detection. Sadeghi et al, 2016~\citep{Sim2RealSadeghi} used domain randomization to decouple different visual environments from the goal of visual servoing. Peng et al, 2017~\citep{DBLP:journals/corr/abs-1710-06537} used domain randomization to enable zero-shot policy transfer from simulation to the real world. Most recently and perhaps most notably, OpenAI, 2018 \citep{openai2018learning} performed domain randomization on the 24 DoF shadow hand~\citep{shadowhand} to transfer a dexterous manipulation task to the real robot.

 Various simulator environments have been released with the aim of creating a common benchmark suite~\citep{dblp:journals/corr/abs-1801-00690, DBLP:journals/corr/abs-1811-02790, openaigym}. In addition there are frameworks based on these popular benchmark libraries to enable more realistic tasks. One such framework is Fan et al's 2018~\citep{pmlr-v87-fan18a} SURREAL environment. The SURREAL framework provides a straightforward framework with practical robotic manipulation tasks, but it is not designed to transfer policies between environments. Hence, we are interested in creating an environment that captures the intersection of domain transfer, practical tasks, and realism.
 



In this paper, we introduce DoorGym, an open-source domain randomized environment for the task of opening various kinds of doors. We also present baseline agents capable of solving this task. We present multiple doorknob types, including round, lever, and pull knobs, with the option of using custom knobs as well. We provide a benchmark environment, and baseline agent capable of opening multiple types of door knobs. We also show that the agent trained in our environment is capable of successfully generalizing to previously unseen environments. 



\section{{DoorGym} Door Opening Task Environment Development}

\begin{figure}[t!]
    \begin{subfigure}[b]{0.24\textwidth}
        \includegraphics[width=\textwidth]{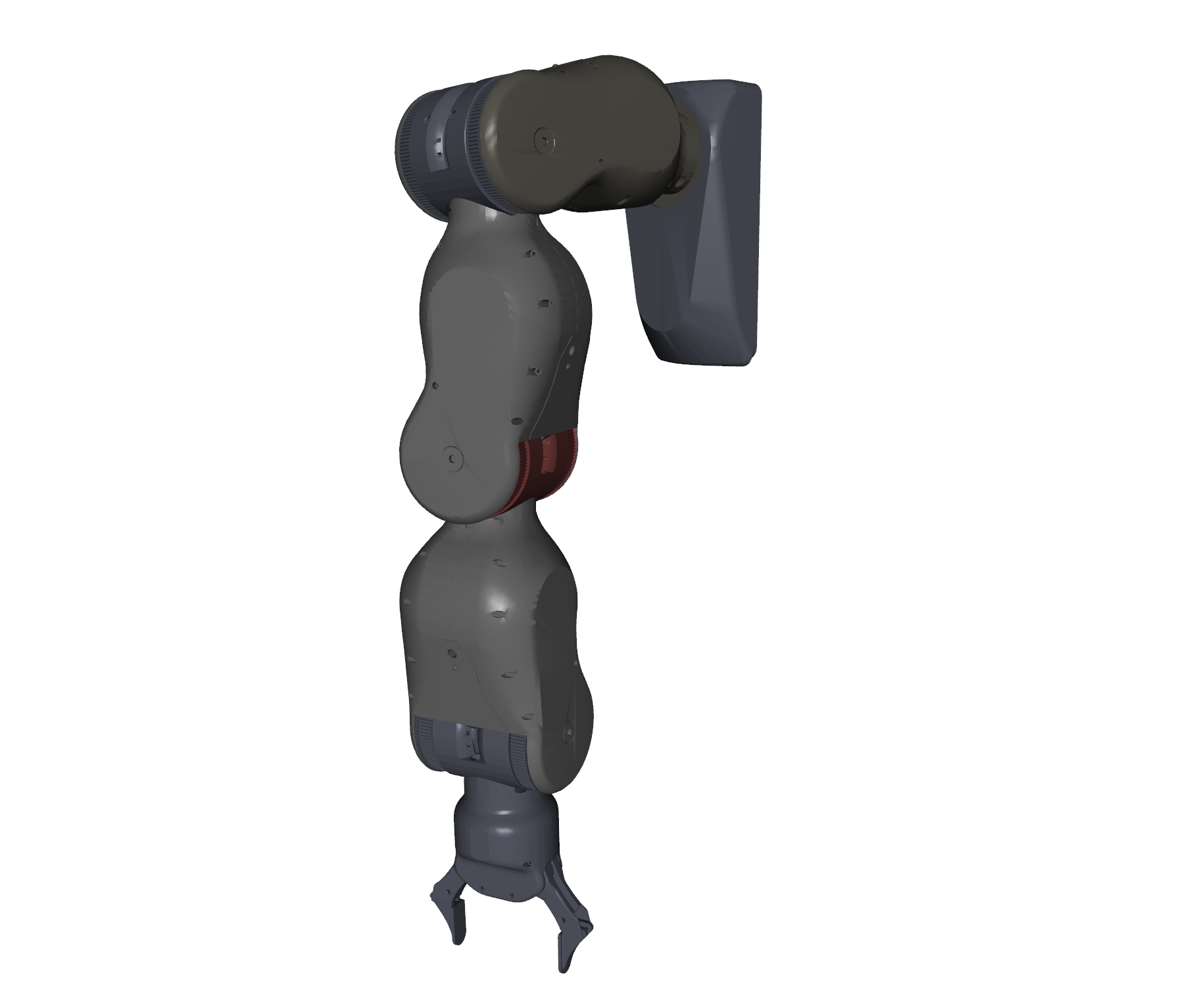}
        \caption{\scriptsize 8DoF BLUE with grippers}
        \label{gripper_types:gripper_1}
    \end{subfigure}
    \begin{subfigure}[b]{0.24\textwidth}
        \includegraphics[width=\textwidth]{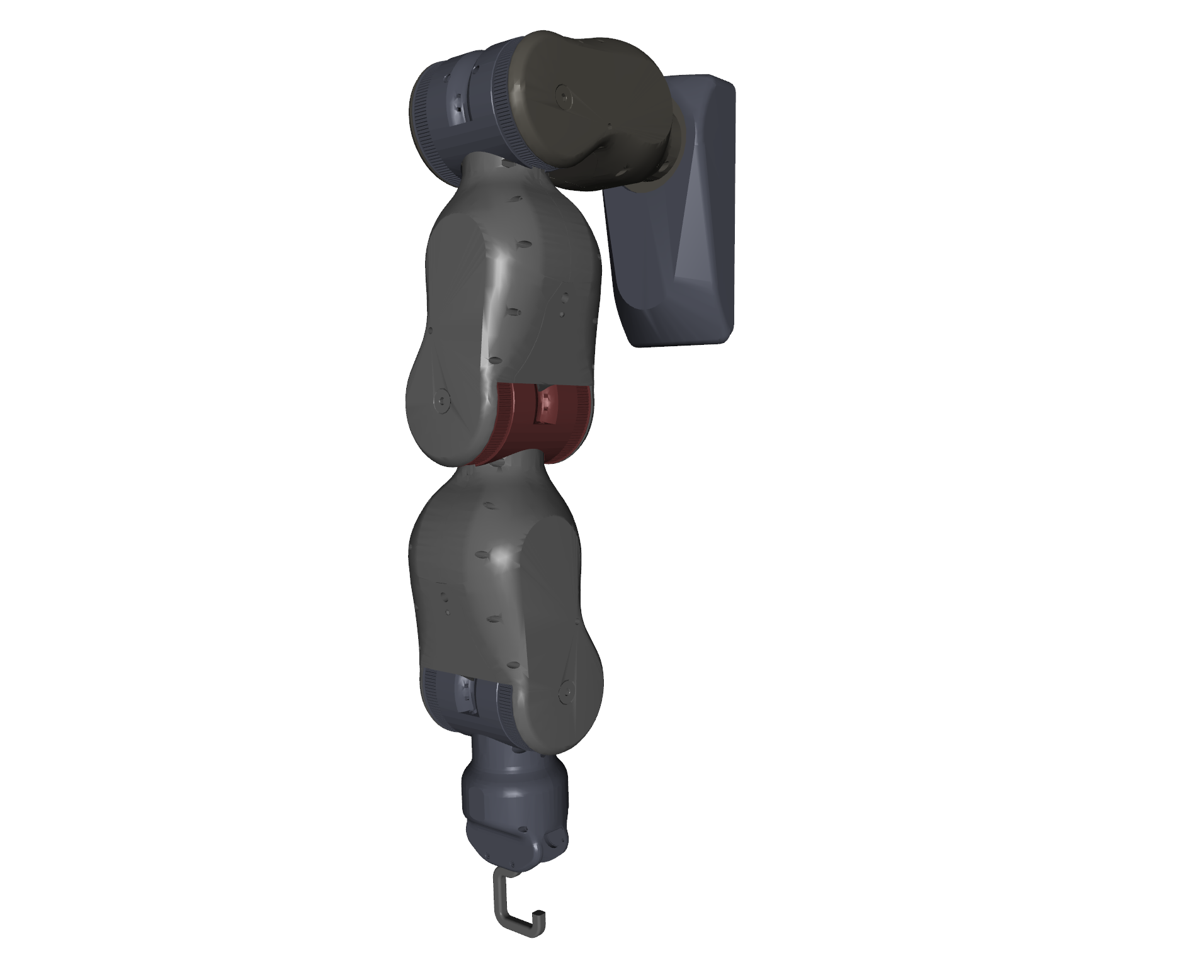}
        \caption{\scriptsize 7DoF BLUE with hook}
        \label{gripper_types:gripper_2}
    \end{subfigure}
    \begin{subfigure}[b]{0.24\textwidth}
        \includegraphics[width=\textwidth]{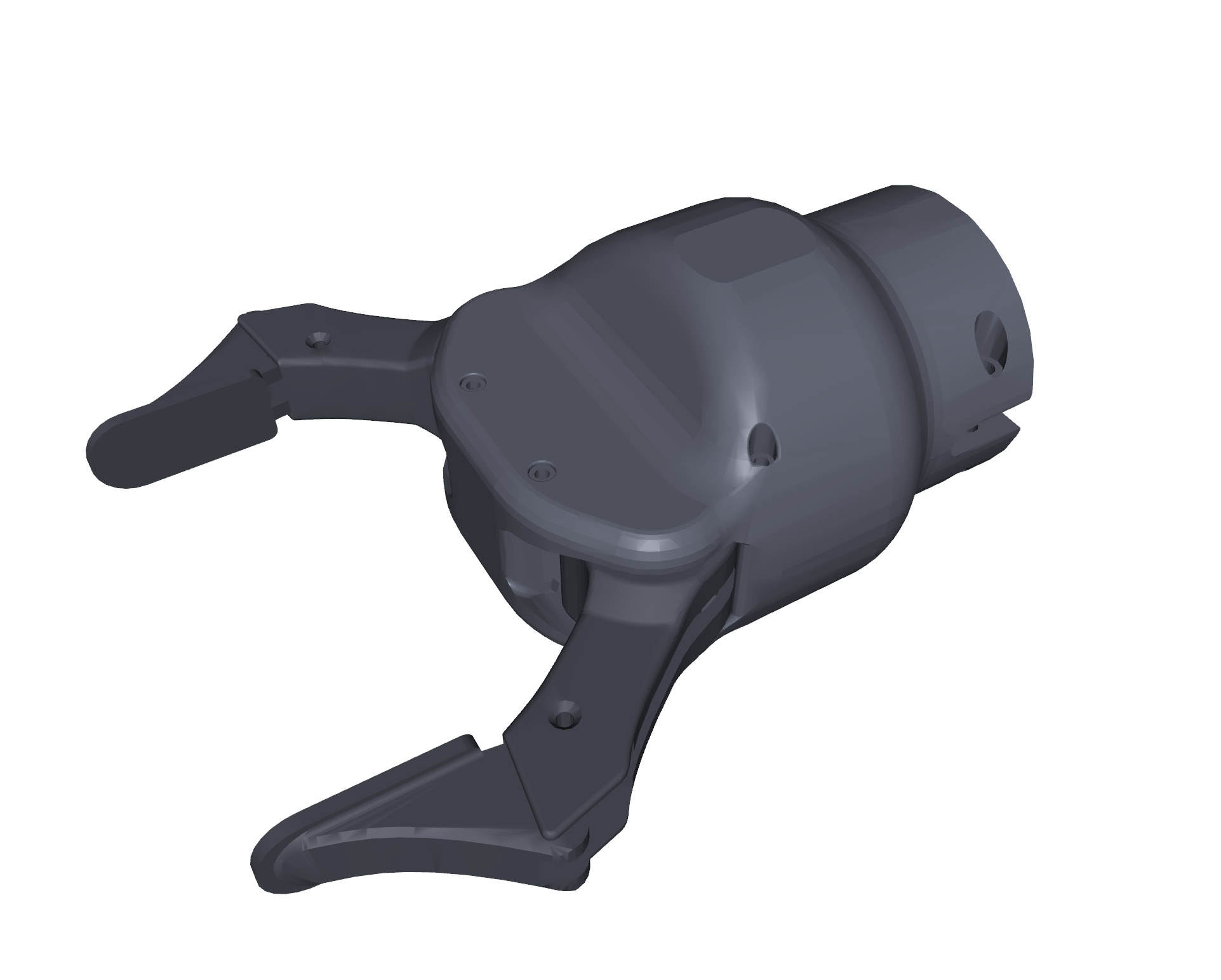}
        \caption{\scriptsize 7DoF Floating gripper}
        \label{gripper_types:gripper_3}
    \end{subfigure}
    \begin{subfigure}[b]{0.24\textwidth}
        \includegraphics[width=\textwidth]{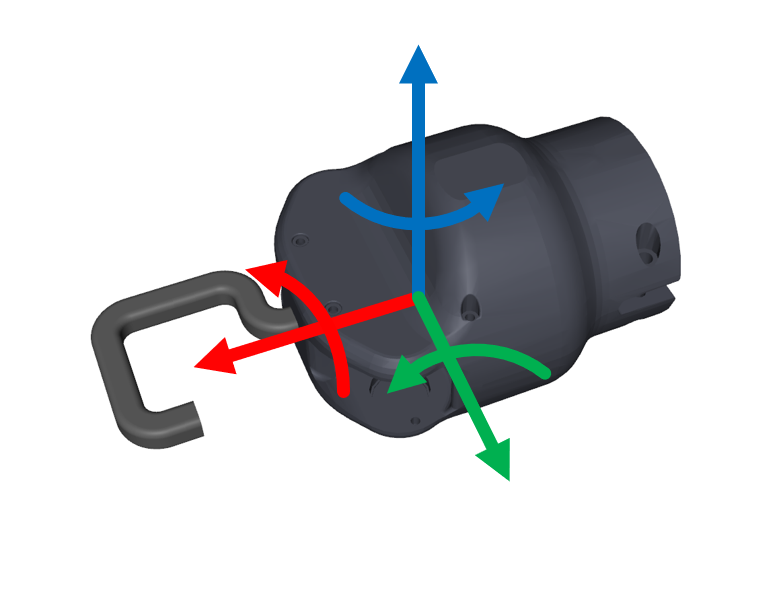}
        \caption{\scriptsize 6DoF Floating hook}
        \label{gripper_types:gripper_4}
    \end{subfigure}
    \caption{Available arm types}
    \label{gripper_types}
\end{figure}

\begin{figure}[t]
    \begin{subfigure}{0.49\textwidth}
        \includegraphics[width=\textwidth]{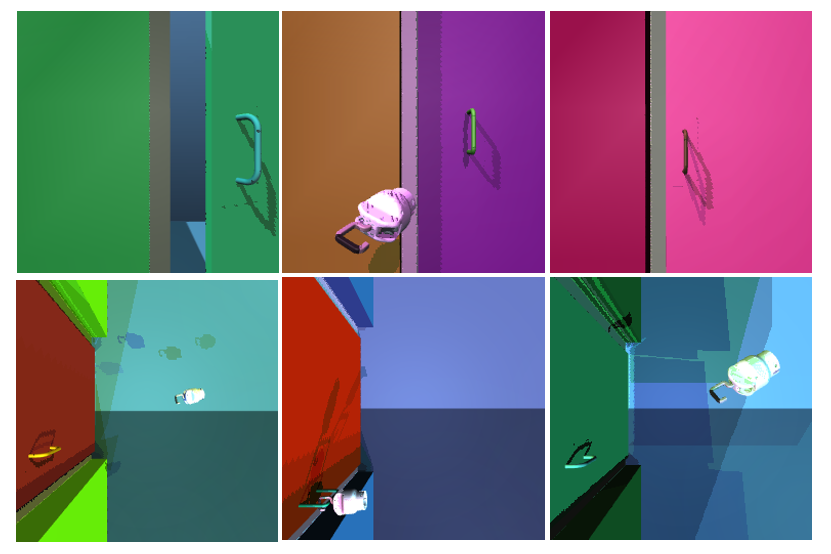}
        \caption{MuJoCo rendered images}
        \label{unity_vs_mujoco:mujoco}
    \end{subfigure}
    \hfill
    \begin{subfigure}{0.49\textwidth}
        \includegraphics[width=\textwidth]{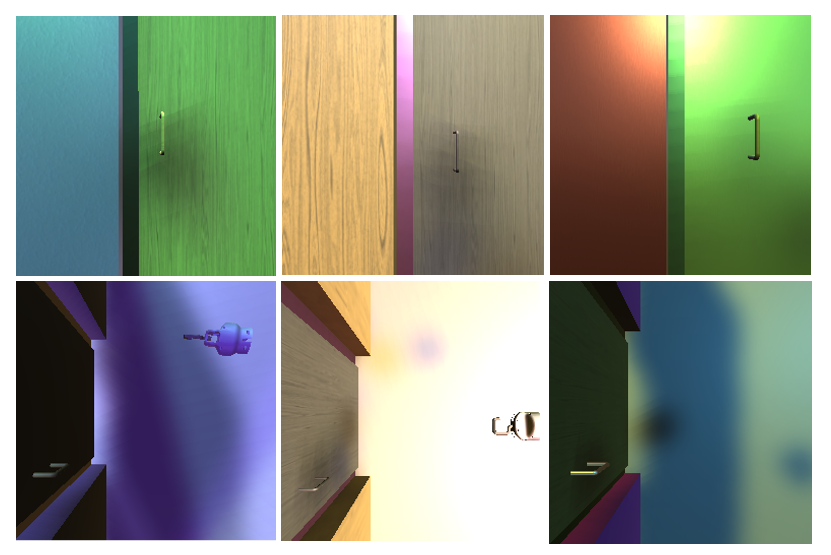}
        \caption{Unity rendered images}
        \label{unity_vs_mujoco:unity}
    \end{subfigure}
    \caption{Comparison between the domain randomized images from MuJoCo and Unity. Unity can render realistic shadows and detailed textures.}
    \label{unity_vs_mujco}
\end{figure}

Our environment is based on the groundwork laid by the OpenAI Gym framework~\citep{openaigym}, and the Unity Game Engine\footnote{Unity game engine website: https://unity3d.com}. Gym offers a variety of discrete as well as continuous environments, including an API for MuJoCo~\citep{6386109}. Combined with wrapper libraries~\citep{dblp:journals/corr/abs-1801-00690, mujocopy}, the MuJoCo API, provides the ability to do elaborate physical simulations as shown in \citep{7989385, openai2018learning}. Our environment makes use of Gym and the MuJoCo API to create an easily randomized world by adding an appropriate doorknob STL file and specifying a configuration. While MuJoCo allows us to create a rich environment, it sacrifices visual realism. For this purpose we use Unity to improve the rendering quality. Our simulator also allows the user to modify the simulator reward and observable state easily. We also provide a MuJoCo-Unity plug-in that allows users to access to photorealistic images and various visual effects easily through a Python API. Figure \ref{unity_vs_mujco} shows the comparison of images by each software.

\subsection{Randomized Door-World Generator}
We provide a simulation environment devkit that includes the dataset of pull knobs, lever knobs, and round knobs, as well as a script to generate more using Autodesk Fusion360~\citep{fusion360}. Each type of doorknob is parameterized to generate unique instances by changing a few attributes in the CAD models.
Sample doorknobs are shown in figure~\ref{doorknob_figure}. The pull knob is the simplest of the included doorknobs. It does not have a latching mechanism; as such, the pull knob environment can be opened by reaching out, grasping, and pulling. Both the lever and round knob are more complex and involve turning the knob to unlatch the door. The round knob environment is the most difficult to open due to the geometry of the knob. Each knob is saved as several STL files with 3D data, a screenshot of the doorknob, and a JSON file with all doorknob metadata. 

We spawn doorknobs in the world randomly using the MuJoCo API. To do this, we generate XML files with all the objects, joints, and physical properties of the world. Though the MuJoCo API supports modifying appearance on the fly, it does not support hot swapping STL models in and out. To deal with this limitation, we generate a new world for every doorknob. The environment itself consists of the robot, door, doorknob, door frame, and wall. All physical properties of the door, doorknob, and robot are randomized in the ranges specified in Table \ref{tab:randomized_doorworld_parameters}. In figure \ref{randomized_environment_figure}, we show examples of randomized environments. At training time, environments will be randomly selected and spawned in the simulator.

\subsection{Robot}
In our environment, we use the Berkeley BLUE Robot arm~\citep{DBLP:journals/corr/abs-1904-03815} by default. We prepare 6 different arm/gripper combinations for BLUE, but provide an interface to add new arms and grippers. The arms that we prepared range from a 6 DoF floating hook to a full 10 DoF arm with gripper and mobile platform. Each arm can be seen in figure \ref{gripper_types}. The mobile platform variants of the BLUE arm are visually identical, but unlock the x-y plane for an additional two degrees of freedom. Arms \ref{gripper_types:gripper_1} and \ref{gripper_types:gripper_2} have their configurations expressed by the joint angles, while the floating wrists \ref{gripper_types:gripper_3} and \ref{gripper_types:gripper_4} are expressed by their x-y-z coordinates and x-y-z rotations.

\section{Benchmark}
We aim to build a simulation environment that can maintain transfer-ability between domains. We use simple and readily available observations from the environment. An observation consists of the position and the velocity of each joint, as well as the position of the door knob and the position of the end-effector in world coordinates. The position of the doorknob in world coordinates can be obtained directly from the simulator, or using a 256x256 RGB image and vision network. 
Policy actions use force for linear actuators and torque for rotational actuators, but can be configured to use position control. The size of the action space corresponds to the DoF of each robot. At training time, we use the following reward function for our baseline agent:

\begin{equation}
    r_t = -a_{0} d_t -a_{1}log(d_t+\alpha) -a_{2}o_t-a_{3}\sqrt{u_t^{2}}+a_{4}\phi_t+a_{5}\psi_t
    \label{shaped_reward_eq}
\end{equation}

where $d_t$ is the distance between the fingertip of the end-effector and the center coordinate of the doorknob, the second term has been added to encourage higher precision when the agent gets close to the target \citep{DBLP:journals/corr/LevineWA15}(Levine et al, 2015), $\alpha$ is set to 0.005, $o_t$ is the difference between the current fingertip orientation of the robot and the ideal orientation to hook/grip the doorknob, $u_t$ is the control input to the system, $\phi_t$ is the angle of the door, and $\psi_t$ is the angle of the door knob. The $t$ subscript indicates the value at time $t$. Weights are set to $a_{0}=1.0$, $a_{1}=1.0$, $a_{2}=1.0$, $a_{3}=1.0$, $a_{4}=30.0$, and $a_{5}=50.0$. When using the pull knob, door knob angle $\psi$ is ignored.

We define a successful attempt as the robot opening the door at least 0.2 rad within 20 seconds. For attempt i, it can be expressed as
\begin{equation}
    \label{indicator_function_eq}
    \mathds{1}_i =
    \begin{cases}
        1 & \textrm{if } \phi > 0.2 \textrm{ rad and t} < 20 \\
        0 & \textrm{otherwise}\\
    \end{cases}
\end{equation}

This gives rise to two potential evaluation metrics. The average success rate of opening a door, and the average time to open a door. We measure both these metrics over 100 attempts.
Average success rate, $r_{ASR}$,  and average time to open a door, $r_{AT}$ can be formalized as 

\begin{equation}
    \label{unshaped_reward_eq}
    r_{\textrm{ASR}} = \frac{1}{100}\sum_{i=1}^{100} \mathds{1}_i, \hphantom{2em} r_{\textrm{AT}}= \frac{1}{n}\sum_{i=1}^{n}t_{i}
\end{equation}

where $t_i$ is the time to completion for successful attempts, $n$ is the number of successful attempts, and the indicator function is defined as in Eq. \ref{indicator_function_eq}.

Our environment currently provides a total of 36 possible environment combinations\footnote{There are 3 grippers, 6 robots, and 2 directions (push/pull)} for evaluation. These environments range in difficulty from the simplest case of a pull knob with floating hook, to the extremely challenging round knob to be opened by BLUE with a gripper.

\section{Door opening agent}
\subsection{Method}
To train our baseline agents, we chose an on-policy and off-policy update algorithm. For the on-policy algorithm, Proximal Policy Optimization (PPO)~\citep{ppo} was chosen due to its high stability. Soft Actor Critic (SAC) ~\citep{DBLP:journals/corr/abs-1801-01290} was chosen as an off-policy algorithm due to its exploration capability and sample efficiency. Details of these algorithms can be found in the appendix in sections \ref{PPO_method} and \ref{SAC_method}.

\subsection{Architecture}
Our agent consists of two networks; a vision network to estimate doorknob position, and a policy network to output actions. The architecture of the vision network is shown in figure \ref{fig:vision_network} in the appendix. We jointly predict the x-y-z location of the doorknob using images from the top and front views. The network performs feature extraction on both views using a  feature extractor network composed of residual blocks with 4x downsampling. The extracted features are then pooled along the channel dimension using global average pooling. The pooled feature maps are then fed into a regression network which is directly optimized for mean squared error. We regularize the network by adding a cross entropy loss between the heatmaps, and the corresponding ground truth of the doorknob location. To balance the regression and heatmap losses we train with a scaling factor of $10^{-3}$ for the heatmap loss. We train the network for 35 epochs using the Adam optimizer with a learning rate of $10^{-3}$ with polynomial learning rate decay and a batch size of 50.

Once we have received as estimate of the doorknob location, we take the difference between our estimate and the end-effector location to produce a direction vector. This direction vector is concatenated with the joint positions and velocities to produce an observation. We then use a network of two fully connected layers with tanh activations to produce the output actions. The structure of our policy network can be seen in figure \ref{fig:policyarch} in the appendix.

\begin{figure}[t!]
    \centering
    \includegraphics[width=1.0\textwidth]{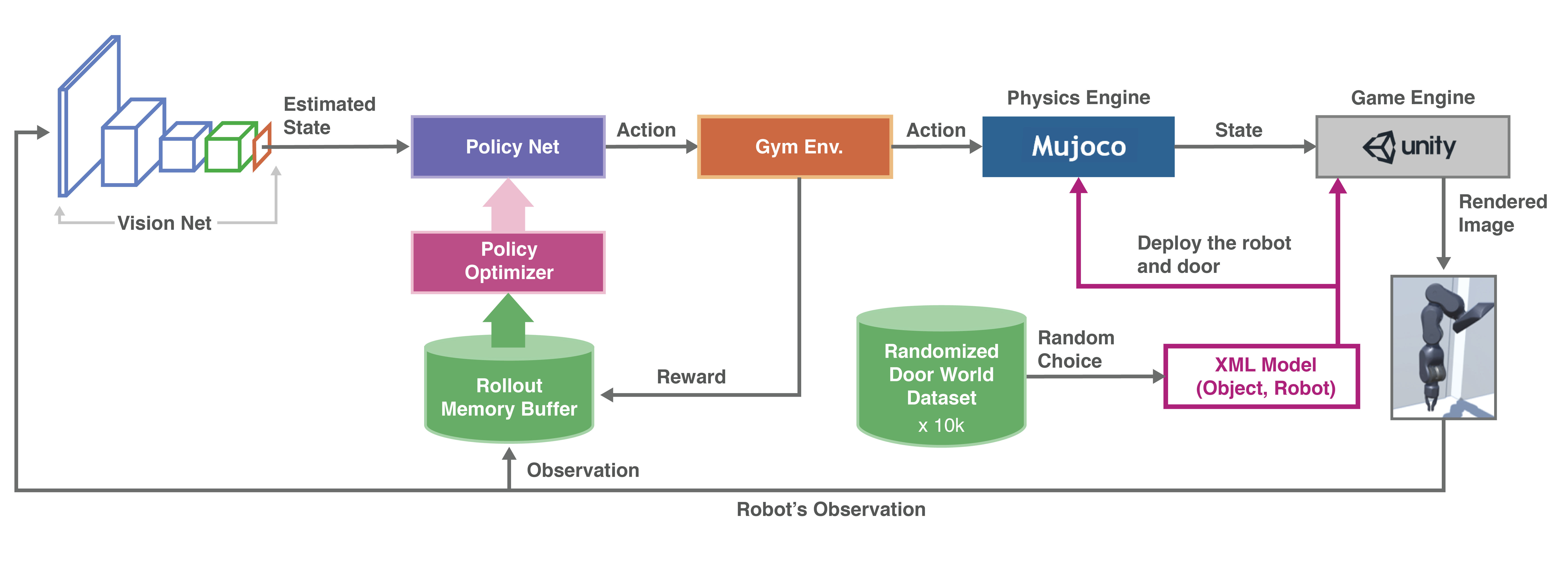}
    \caption{Training Pipeline}
    \label{training_pipeline}
\end{figure}
\subsection{Training Pipeline}
\label{sec:pipeline}

The entire training flow is shown in figure \ref{training_pipeline}. A door-world is chosen randomly from the door-world data set and spawned into MuJoCo simulator and Unity. The policy network decides the next action given the observation of the environment. The Gym environment then returns a reward and an observation of the environment for the next time step. During rollouts, observations, actions and rewards are stored in the rollout memory buffer and used for a policy update.

In order to update the policy efficiently while using the randomized data, PPO method runs 8 workers synchronously. During the training time, door-worlds are randomly sampled and fed to each worker after every update. At the start of each episode, the robot position and orientation are randomly initialized. 
Each episode is 512 time steps (10.2 seconds). The rollout memory buffer of each worker stores 8 episodes for every policy update. During the policy update, rollout memory buffers of all 8 workers are gathered and will be used to update the shared worker policy. We used a learning rate of $10^{-3}$ for all environments. This PPO training pipeline was structured based on code by Kostrikov, 2018~\citep{pytorchrl}. In contrast, due to the fact that SAC is an off-policy RL algorithm, it is trained in different way. For each epoch, a single worker performs 10 episodes in randomly sampled door-worlds and stores this data in the rollout memory buffers which has a maximum size of $10^6$ time steps. When the policy of worker is updated, an accumulated data in the rollout memory buffers will be used. This training pipeline was structured based on rlkit\footnote{https://github.com/vitchyr/rlkit}.

For PPO training, in order to accelerate the training speed and make the policy converge stably, the policies used with the vision network were pretrained using the door knob position from the simulator, followed a training using the door knob position from the simulator with gaussian noise. The gaussian noise used for pretraining approximates the error introduced by the vision network's position estimate. The training using vision network for SAC was done without the pretraining process. All policy training hyperparameters are listed in table \ref{tab:PPO_hyperparameters} and table \ref{tab:SAC_hyperparameters} in the appendix.

\subsection{Experiments}
\label{sec:experiment}
The following three experiments are conducted to evaluate performance as a representative of the environments. Task 1 has a pull knob environment with floating hook, task 2 has a lever knob with floating hook, and task 3 has a pull knob with the BLUE-with-gripper platform. All experiments are evaluated using the unshaped reward as defined in section 4.1. For each task, we have prepared 100 test door-worlds. The agent has 10.2 seconds to try open the door for each world. 

\begin{figure}[]
    \centering
    \includegraphics[width=0.75\textwidth]{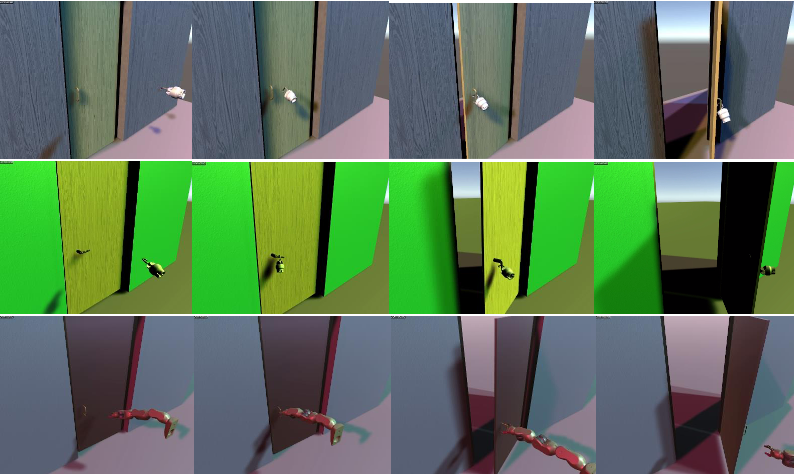}
    \caption{Behavior of successful policies in Task 1, Task 2, and Task 3 respectively.}
    \label{result_pic}
\end{figure}

Table~\ref{benchmark_result} shows the results of each task using average success rate and average opening time for different door knob position estimation methods. Each row corresponds to a different knob position estimation method. By using the ground truth of the door knob position from the simulator, the agent has a success rate of over 70\% for PPO. In contrast, SAC generally has lower success rates, and specifically it has a 0\% success rate for opening the lever knob. During training, SAC shows better exploration capability and its total reward converges faster than PPO, but the result suggests that PPO has better exploitation capabilities than SAC in trade-off of its training speed. PPO's success rate of all tasks decreases after adding gaussian noise to the door knob position. Tasks that require accurate position information to execute complicated manipulation such as the lever knob suffer more than other tasks. When the door knob position information comes from the vision network, the success rates of both algorithm decrease even more. This result implies that position estimation in the 3D space of the door knob is extremely important for the door opening task. Our vision network achieves $\pm$3.079 cm accuracy as shown in table \ref{vision_net_results_tab} in the appendix. Even with this level of accuracy, it was not enough to adapt to open the randomized door. Moreover, even though we train the vision network with the images that has randomly located robot, at the inference time, it is possible that robot is located in right in front of the camera and lessens the accuracy of the position estimation. Figure \ref{result_pic} shows the behavior of the robot in each task in the successful case. 
Since the door angle reward is weighted more heavily than other rewards, the robot tries to open the door after it hooks or rotates the door knob. Results of all combinations of the environments and the algorithms are shown in table \ref{tab:full_result_PPO}, \ref{tab:full_result_SAC} in the appendix.

\begin{table}[]
\caption{Average Success Ratio and Average Time to Open measured in seconds.}
    \label{benchmark_result}
    \centering
\begin{tabular}{llllllll}
\cline{2-8}
 &  & \multicolumn{2}{c}{Task 1} & \multicolumn{2}{c}{Task 2} & \multicolumn{2}{c}{Task 3} \\ \cline{3-8} 
 &  & $r_{\textrm{ASR}}$ & $r_{\textrm{AT}}$ & $r_{\textrm{ASR}}$ & $r_{\textrm{AT}}$ & $r_{\textrm{ASR}}$ & $r_{\textrm{AT}}$ \\ \hline
 & Ground Truth Position & 0.95 & 4.51 & 0.68 & 4.12 & 0.71 & 5.13 \\
PPO & Ground Truth Position + N(0, $\sigma$) & 0.78 & 5.18 & 0.54 & 4.11 & 0.65 & 4.73 \\
 & Vision Network Estimated Position & 0.48 & 6.15 & 0.29 & 3.83 & 0.57 & 5.12 \\ \hline
 & Ground Truth Position & 0.23 & 4.29 & 0 & N/A & 0.33 & 5.02 \\
SAC & Ground Truth Position + N(0, $\sigma$) & - & - & - & - & - & - \\
 & Vision Network Estimated Position & 0 & N/A & 0 & N/A & 0.05 & 5.06 \\ \hline
\end{tabular}
\end{table}

\begin{table}[]
    \caption{Success ratio of a policy trained only on a single environment (env1)}
    \label{ablation_policy_single_env}
     \makebox[\linewidth]{
    
    \begin{tabular}{ccccccccccccc}
    \hline
    \multicolumn{1}{l}{} & \multicolumn{6}{c|}{Trained on env1} & \multicolumn{6}{c}{Trained on Randomized env.} \\ \cline{2-13} 
    Test Environment & \multicolumn{2}{c}{GT} & \multicolumn{2}{c}{GT + Noise} & \multicolumn{2}{c|}{Vision} & \multicolumn{2}{c}{GT} & \multicolumn{2}{c}{GT + Noise} & \multicolumn{2}{c}{Vision} \\ \cline{2-13} 
     & \multicolumn{1}{l}{$r_{ASR}$} & $r_{AT}$ & \multicolumn{1}{l}{$r_{ASR}$} & $r_{AT}$ & \multicolumn{1}{l}{$r_{ASR}$} & \multicolumn{1}{c|}{$r_{AT}$} & $r_{ASR}$ & $r_{AT}$ & $r_{ASR}$ & $r_{AT}$ & $r_{ASR}$ & $r_{AT}$ \\ \hline
    Tested on env1 & 0.92 & 4.71 & 0.84 & 4.91 & 0.46 & \multicolumn{1}{c|}{6.93} & 0.99 & 3.78 & 0.87 & 4.39 & 0.50 & 6.74 \\ \hline
    Tested on Randomized env. & 0.50 & 4.47 & 0.52 & 7.13 & 0.0 & \multicolumn{1}{c|}{-} & 0.95 & 4.51 & 0.78 & 5.18 & 0.48 & 6.15 \\ \hline
    \end{tabular}
    }
\end{table}

\subsection{Ablation Study}
We performed additional experiments to confirm the necessity of DR for policy transferability, and also we show that we were able to transfer our vision network from simulation to real doorknobs. Table 1 below shows the ablation test comparing policies trained with DR and without DR using PPO algorithm. On the left side of Table \ref{ablation_policy_single_env} shows the agent trained in a single environment (env1) achieves 92\% of success rate in that environment, but it struggles in randomized environments (50\% success). It has a similar trend using the vision network for position estimation (From 46\% to 0\%). 

On the right side of Table \ref{ablation_policy_single_env}, the agents are trained under randomized environment and it is showing its robust performance. The agent trained in randomized environments have high success rates in both env1(99\%) and randomized env(95\%) with the ground truth doorknob position. By conducting the same test using the vision network estimated doorknob position, the randomized trained policy have almost the same performance no matter the test environments are randomized or fixed parameters environments (50\% and 48\%).

For the vision network, we collected and annotated 20 round knob images in homes and performed inference on the vision network. The network trained with DR achieved an accuracy of 4.95cm in the real world, whereas the network trained without DR performs significantly worse at 19.53cm. Details of these results can be found in appendix chapter \ref{Vision Network Results}. From these results we can say that DR is required to make both the vision network and policy networks robust enough to work in different domains. 

\section{Real World Transfer}
\begin{figure}[t!]
    \begin{subfigure}[b]{0.45\textwidth}
        \centering
        \includegraphics[width=0.75\textwidth]{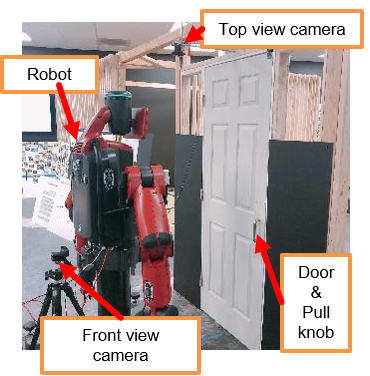}
        \caption{Experiment Appearance}
        \label{appearance}
    \end{subfigure}
    \begin{subfigure}[b]{0.45\textwidth}
        \centering
        \includegraphics[width=0.9\textwidth]{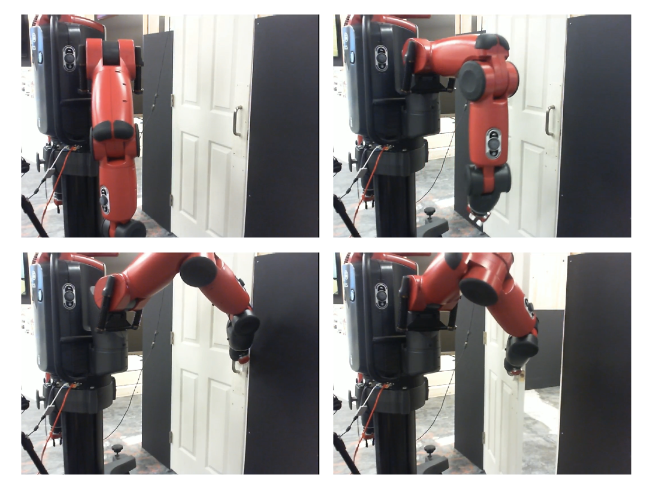}
        \caption{Sequential snapshot of door-opening motion}
        \label{cameraview}
    \end{subfigure}
    \caption{Real world experiment}
    \label{realworld}
\end{figure}

In order to check the robustness of the trained policy, we performed zero-shot sim-to-real policy transfer. The experiment was executed using the Baxter robot and a pull knob equipped wooden door (Height: 2100mm, Width:820mm, Weight: 11.3kg). The setup is shown in Figure \ref{appearance}. The front view camera, top view camera, and robot are placed at the same coordinates as in the simulator. During the experiment, only the right arm of the robot is active since the doorknob is placed on the right side of the door, and the arm starts from the hanging position. The policy used in the experiment is trained purely in simulation using PPO with same hyperparameters as in section \ref{sec:experiment}. The location of the doorknob is estimated by the vision network. The control loop frequency is 50Hz. 

The experiments were evaluated 100 times, with an experiment episode length of 20.0s. If the robot opened the door more than 0.2 rad during the episode it is counted as a success. The results of using Baxter in both simulation and the real world are shown in Table \ref{realworld_result}. Result shows that even though the success rate decrease from 70\% to 59\%, the policy is able to open the door in the real world. However, the behavior of the robot looks unstable, and also the total time to open the door is significantly longer than in simulation.

\begin{table}[h]
\caption{Results of Real World Transfer Experiment}
    \centering
    \begin{tabular}{@{}lll@{}}
    \toprule
              & $r_{ASR}$ & $r_{AT}$ \\ \midrule
    Simulator  & 0.70   & 4.71  \\
    Real World & 0.59   & 16.25 \\ \bottomrule          
    \end{tabular}
    \label{realworld_result}
\end{table}




\section{Conclusion}

We presented DoorGym, a simulator environment that supports many degrees of domain randomization on a door opening task with varying degrees of difficulty. As a starting point, we presented a baseline agent based on PPO and SAC, which is capable of opening doors in novel environments. We evaluated the success of this agent by measuring the success rate in 100 attempts, with a success rate of up to 95\%. Moreover, we were able to transfer a trained policy from simulation to the real world. Future work will involve expanding the baseline networks as well as incorporating more complicated tasks such as a broader range of doorknobs, locked doors, door knob generalization, and multi-agent scenarios.


\pagebreak
\bibliography{neurips_2019}

\begin{thebibliography}{34}
\providecommand{\natexlab}[1]{#1}
\providecommand{\url}[1]{\texttt{#1}}
\expandafter\ifx\csname urlstyle\endcsname\relax
  \providecommand{\doi}[1]{doi: #1}\else
  \providecommand{\doi}{doi: \begingroup \urlstyle{rm}\Url}\fi

\bibitem[{Klingbeil} et~al.(2010){Klingbeil}, {Saxena}, and {Ng}]{5649847}
E.~{Klingbeil}, A.~{Saxena}, and A.~Y. {Ng}.
\newblock Learning to open new doors.
\newblock In \emph{2010 IEEE/RSJ International Conference on Intelligent Robots
  and Systems}, pages 2751--2757, 2010.

\bibitem[Du et~al.(2015)Du, Feng, Franklin, Gennert, Graff, He, Jaeger, Kim,
  Knoedler, Li, Liu, Long, Padir, Polido, Tighe, and Xinjilefu]{du2015whatha}
R.~Du, S.~Feng, P.~Franklin, M.~Gennert, J.~P. Graff, P.~He, A.~Jaeger, J.~Kim,
  K.~Knoedler, L.~Li, C.~Y. Liu, X.~Long, T.~Padir, F.~Polido, G.~G. Tighe, and
  X.~Xinjilefu.
\newblock What happened at the darpa robotics challenge , and why ?
\newblock In \emph{None}, 2015.

\bibitem[{Nemec} et~al.(2017){Nemec}, {Žlajpah}, and {Ude}]{8023522}
B.~{Nemec}, L.~{Žlajpah}, and A.~{Ude}.
\newblock Door opening by joining reinforcement learning and intelligent
  control.
\newblock In \emph{2017 18th International Conference on Advanced Robotics
  (ICAR)}, 2017.

\bibitem[{Gu} et~al.(2017){Gu}, {Holly}, {Lillicrap}, and {Levine}]{7989385}
S.~{Gu}, E.~{Holly}, T.~{Lillicrap}, and S.~{Levine}.
\newblock Deep reinforcement learning for robotic manipulation with
  asynchronous off-policy updates.
\newblock In \emph{2017 IEEE International Conference on Robotics and
  Automation (ICRA)}, pages 3389--3396, May 2017.
\newblock \doi{10.1109/ICRA.2017.7989385}.

\bibitem[{Karayiannidis} et~al.(2012){Karayiannidis}, {Smith}, {Viña},
  {Ogren}, and {Kragic}]{6385835}
Y.~{Karayiannidis}, C.~{Smith}, F.~E. {Viña}, P.~{Ogren}, and D.~{Kragic}.
\newblock “open sesame!” adaptive force/velocity control for opening
  unknown doors.
\newblock In \emph{2012 IEEE/RSJ International Conference on Intelligent Robots
  and Systems}, 2012.

\bibitem[{Endres} et~al.(2013){Endres}, {Trinkle}, and {Burgard}]{endres2013}
F.~{Endres}, J.~{Trinkle}, and W.~{Burgard}.
\newblock Learning the dynamics of doors for robotic manipulation.
\newblock In \emph{2013 IEEE/RSJ International Conference on Intelligent Robots
  and Systems}, pages 3543--3549, 2013.

\bibitem[{Anguelov} et~al.(2004){Anguelov}, {Koller}, {Parker}, and
  {Thrun}]{1308857}
D.~{Anguelov}, D.~{Koller}, E.~{Parker}, and S.~{Thrun}.
\newblock Detecting and modeling doors with mobile robots.
\newblock In \emph{IEEE International Conference on Robotics and Automation,
  2004. Proceedings. ICRA '04. 2004}, volume~4, pages 3777--3784 Vol.4, April
  2004.
\newblock \doi{10.1109/ROBOT.2004.1308857}.

\bibitem[{Rusu} et~al.(2009){Rusu}, {Meeussen}, {Chitta}, and {Beetz}]{5174717}
R.~B. {Rusu}, W.~{Meeussen}, S.~{Chitta}, and M.~{Beetz}.
\newblock Laser-based perception for door and handle identification.
\newblock In \emph{2009 International Conference on Advanced Robotics}, pages
  1--8, June 2009.

\bibitem[{Kalakrishnan} et~al.(2011){Kalakrishnan}, {Righetti}, {Pastor}, and
  {Schaal}]{6095096}
M.~{Kalakrishnan}, L.~{Righetti}, P.~{Pastor}, and S.~{Schaal}.
\newblock Learning force control policies for compliant manipulation.
\newblock In \emph{2011 IEEE/RSJ International Conference on Intelligent Robots
  and Systems}, pages 4639--4644, Sep. 2011.
\newblock \doi{10.1109/IROS.2011.6095096}.

\bibitem[Theodorou et~al.(2010)Theodorou, Buchli, and
  Schaal]{Theodorou:2010:GPI:1756006.1953033}
E.~Theodorou, J.~Buchli, and S.~Schaal.
\newblock A generalized path integral control approach to reinforcement
  learning.
\newblock \emph{J. Mach. Learn. Res.}, 11:\penalty0 3137--3181, Dec. 2010.
\newblock ISSN 1532-4435.

\bibitem[Rajeswaran et~al.(2017)Rajeswaran, Kumar, Gupta, Schulman, Todorov,
  and Levine]{DBLP:journals/corr/abs-1709-10087}
A.~Rajeswaran, V.~Kumar, A.~Gupta, J.~Schulman, E.~Todorov, and S.~Levine.
\newblock Learning complex dexterous manipulation with deep reinforcement
  learning and demonstrations.
\newblock \emph{CoRR}, abs/1709.10087, 2017.
\newblock URL \url{http://arxiv.org/abs/1709.10087}.

\bibitem[Rajeswaran et~al.(2016)Rajeswaran, Ghotra, Levine, and
  Ravindran]{DBLP:journals/corr/RajeswaranGLR16}
A.~Rajeswaran, S.~Ghotra, S.~Levine, and B.~Ravindran.
\newblock Epopt: Learning robust neural network policies using model ensembles.
\newblock \emph{CoRR}, abs/1610.01283, 2016.
\newblock URL \url{http://arxiv.org/abs/1610.01283}.

\bibitem[Sadeghi and Levine(2016)]{CAD2RLSadeghi}
F.~Sadeghi and S.~Levine.
\newblock (cad){\textdollar}{\^{}}2{\textdollar}rl: Real single-image flight
  without a single real image.
\newblock \emph{CoRR}, abs/1611.04201, 2016.
\newblock URL \url{http://arxiv.org/abs/1611.04201}.

\bibitem[Tobin et~al.(2017)Tobin, Fong, Ray, Schneider, Zaremba, and
  Abbeel]{DBLP:journals/corr/TobinFRSZA17}
J.~Tobin, R.~Fong, A.~Ray, J.~Schneider, W.~Zaremba, and P.~Abbeel.
\newblock Domain randomization for transferring deep neural networks from
  simulation to the real world.
\newblock \emph{CoRR}, abs/1703.06907, 2017.

\bibitem[Sadeghi et~al.(2017)Sadeghi, Toshev, Jang, and
  Levine]{Sim2RealSadeghi}
F.~Sadeghi, A.~Toshev, E.~Jang, and S.~Levine.
\newblock Sim2real view invariant visual servoing by recurrent control.
\newblock \emph{CoRR}, abs/1712.07642, 2017.
\newblock URL \url{http://arxiv.org/abs/1712.07642}.

\bibitem[Peng et~al.(2017)Peng, Andrychowicz, Zaremba, and
  Abbeel]{DBLP:journals/corr/abs-1710-06537}
X.~B. Peng, M.~Andrychowicz, W.~Zaremba, and P.~Abbeel.
\newblock Sim-to-real transfer of robotic control with dynamics randomization.
\newblock \emph{CoRR}, abs/1710.06537, 2017.

\bibitem[OpenAI et~al.(2018)OpenAI, Andrychowicz, Baker, Chociej, Józefowicz,
  McGrew, Pachocki, Petron, Plappert, Powell, Ray, Schneider, Sidor, Tobin,
  Welinder, Weng, and Zaremba]{openai2018learning}
OpenAI, M.~Andrychowicz, B.~Baker, M.~Chociej, R.~Józefowicz, B.~McGrew,
  J.~Pachocki, A.~Petron, M.~Plappert, G.~Powell, A.~Ray, J.~Schneider,
  S.~Sidor, J.~Tobin, P.~Welinder, L.~Weng, and W.~Zaremba.
\newblock Learning dexterous in-hand manipulation.
\newblock \emph{CoRR}, 2018.
\newblock URL \url{http://arxiv.org/abs/1808.00177}.

\bibitem[ShadowRobot(2005)]{shadowhand}
ShadowRobot.
\newblock Shadowrobot dexterous hand, 2005.
\newblock URL \url{https://www.shadowrobot.com/products/dexterous-hand/}.

\bibitem[Tassa et~al.(2018)Tassa, Doron, Muldal, Erez, Li, de~Las~Casas,
  Budden, Abdolmaleki, Merel, Lefrancq, Lillicrap, and
  Riedmiller]{dblp:journals/corr/abs-1801-00690}
Y.~Tassa, Y.~Doron, A.~Muldal, T.~Erez, Y.~Li, D.~de~Las~Casas, D.~Budden,
  A.~Abdolmaleki, J.~Merel, A.~Lefrancq, T.~P. Lillicrap, and M.~A. Riedmiller.
\newblock Deepmind control suite.
\newblock \emph{CoRR}, abs/1801.00690, 2018.
\newblock URL \url{http://arxiv.org/abs/1801.00690}.

\bibitem[Mandlekar et~al.(2018)Mandlekar, Zhu, Garg, Booher, Spero, Tung, Gao,
  Emmons, Gupta, Orbay, Savarese, and
  Fei{-}Fei]{DBLP:journals/corr/abs-1811-02790}
A.~Mandlekar, Y.~Zhu, A.~Garg, J.~Booher, M.~Spero, A.~Tung, J.~Gao, J.~Emmons,
  A.~Gupta, E.~Orbay, S.~Savarese, and L.~Fei{-}Fei.
\newblock Roboturk: {A} crowdsourcing platform for robotic skill learning
  through imitation.
\newblock \emph{CoRR}, abs/1811.02790, 2018.
\newblock URL \url{bit.ly/2XJsT9N}.

\bibitem[Brockman et~al.(2016)Brockman, Cheung, Pettersson, Schneider,
  Schulman, Tang, and Zaremba]{openaigym}
G.~Brockman, V.~Cheung, L.~Pettersson, J.~Schneider, J.~Schulman, J.~Tang, and
  W.~Zaremba.
\newblock Openai gym, 2016.

\bibitem[Fan et~al.(2018)Fan, Zhu, Zhu, Liu, Zeng, Gupta, Creus-Costa,
  Savarese, and Fei-Fei]{pmlr-v87-fan18a}
L.~Fan, Y.~Zhu, J.~Zhu, Z.~Liu, O.~Zeng, A.~Gupta, J.~Creus-Costa, S.~Savarese,
  and L.~Fei-Fei.
\newblock Surreal: Open-source reinforcement learning framework and robot
  manipulation benchmark.
\newblock In \emph{Proceedings of The 2nd Conference on Robot Learning},
  volume~87 of \emph{Proceedings of Machine Learning Research}, pages 767--782.
  PMLR, 29--31 Oct 2018.
\newblock URL \url{http://proceedings.mlr.press/v87/fan18a.html}.

\bibitem[{Todorov} et~al.(2012){Todorov}, {Erez}, and {Tassa}]{6386109}
E.~{Todorov}, T.~{Erez}, and Y.~{Tassa}.
\newblock Mujoco: A physics engine for model-based control.
\newblock In \emph{2012 IEEE/RSJ International Conference on Intelligent Robots
  and Systems}, pages 5026--5033, 2012.

\bibitem[OpenAI(2018)]{mujocopy}
OpenAI.
\newblock Mujoco-py.
\newblock \url{https://github.com/openai/mujoco-py}, 2018.

\bibitem[{AutoDesk, Inc.}(2013)]{fusion360}
{AutoDesk, Inc.}
\newblock Autodesk fusion360, 2013.
\newblock URL \url{https://autode.sk/2XvQgiL}.

\bibitem[Gealy et~al.(2019)Gealy, McKinley, Yi, Wu, Downey, Balke, Zhao, Guo,
  Thomasson, Sinclair, Cuellar, McCarthy, and
  Abbeel]{DBLP:journals/corr/abs-1904-03815}
D.~V. Gealy, S.~McKinley, B.~Yi, P.~Wu, P.~R. Downey, G.~Balke, A.~Zhao,
  M.~Guo, R.~Thomasson, A.~Sinclair, P.~Cuellar, Z.~McCarthy, and P.~Abbeel.
\newblock Quasi-direct drive for low-cost compliant robotic manipulation.
\newblock \emph{CoRR}, abs/1904.03815, 2019.
\newblock URL \url{http://arxiv.org/abs/1904.03815}.

\bibitem[Levine et~al.(2015)Levine, Wagener, and
  Abbeel]{DBLP:journals/corr/LevineWA15}
S.~Levine, N.~Wagener, and P.~Abbeel.
\newblock Learning contact-rich manipulation skills with guided policy search.
\newblock \emph{CoRR}, abs/1501.05611, 2015.
\newblock URL \url{http://arxiv.org/abs/1501.05611}.

\bibitem[John~Schulman(2017)]{ppo}
P.~D. A. R. O.~K. John~Schulman, Filip~Wolski.
\newblock Proximal policy optimization algorithms.
\newblock \emph{CoRR}, 2017.
\newblock URL \url{https://arxiv.org/abs/1707.06347}.

\bibitem[Haarnoja et~al.(2018)Haarnoja, Zhou, Abbeel, and
  Levine]{DBLP:journals/corr/abs-1801-01290}
T.~Haarnoja, A.~Zhou, P.~Abbeel, and S.~Levine.
\newblock Soft actor-critic: Off-policy maximum entropy deep reinforcement
  learning with a stochastic actor.
\newblock \emph{CoRR}, abs/1801.01290, 2018.

\bibitem[Kostrikov(2018)]{pytorchrl}
I.~Kostrikov.
\newblock Pytorch implementations of reinforcement learning algorithms.
\newblock \url{https://github.com/ikostrikov/pytorch-a2c-ppo-acktr-gail}, 2018.

\bibitem[Williams(1992)]{Williams1992}
R.~J. Williams.
\newblock Simple statistical gradient-following algorithms for connectionist
  reinforcement learning.
\newblock \emph{Machine Learning}, 8\penalty0 (3):\penalty0 229--256, 1992.
\newblock ISSN 1573-0565.
\newblock \doi{10.1007/BF00992696}.

\bibitem[Schulman et~al.(2016)Schulman, Moritz, Levine, Jordan, and
  Abbeel]{Schulmanetal_ICLR2016}
J.~Schulman, P.~Moritz, S.~Levine, M.~Jordan, and P.~Abbeel.
\newblock High-dimensional continuous control using generalized advantage
  estimation.
\newblock In \emph{Proceedings of the International Conference on Learning
  Representations (ICLR)}, 2016.

\bibitem[Haarnoja et~al.(2018)Haarnoja, Zhou, Hartikainen, Tucker, Ha, Tan,
  Kumar, Zhu, Gupta, Abbeel, and Levine]{DBLP:journals/corr/abs-1812-05905}
T.~Haarnoja, A.~Zhou, K.~Hartikainen, G.~Tucker, S.~Ha, J.~Tan, V.~Kumar,
  H.~Zhu, A.~Gupta, P.~Abbeel, and S.~Levine.
\newblock Soft actor-critic algorithms and applications.
\newblock \emph{CoRR}, abs/1812.05905, 2018.
\newblock URL \url{http://arxiv.org/abs/1812.05905}.

\bibitem[Kingma and Ba(2014)]{Adam}
D.~P. Kingma and J.~Ba.
\newblock Adam: A method for stochastic optimization, 2014.
\newblock cite arxiv:1412.6980Comment: Published as a conference paper at the
  3rd International Conference for Learning Representations, San Diego, 2015.

\end{thebibliography}
\pagebreak

\section{Appendix}


\subsection{Proximal Policy Optimization (PPO)}
\label{PPO_method}
Since the door opening task is a complex continuous control problem, a policy gradient method~\citep{Williams1992} (Williams et al, 1992) is one of a potent method to use. Usually, direct policy search suffers from instability due to high gradient variance. While this can be mitigated by using a larger batch size during training, policy gradient algorithms can be unstable. Proximal Policy Optimization (PPO) is a policy gradient method that deals with this instability by constraining the objective function, 

\begin{equation} \label{eq:1}
    \mathcal{L}_{\mathrm{PPO}}=\left[ \min \left(\frac{\pi\left(a_{t} | s_{t}\right)}{\pi_{\mathrm{old}}\left(a_{t} | s_{t}\right)} \hat{A}_{t}, \operatorname{clip}\left(\frac{\pi\left(a_{t} | s_{t}\right)}{\pi_{\mathrm{old}}\left(a_{t} | s_{t}\right)}, 1-\epsilon, 1+\epsilon\right) \hat{A}_{t}\right)\right]
\end{equation}
where $\pi$ and $\pi_{\textrm{old}}$ are the current and previous policies respectively, $\hat{A}_t$ is the advantage and $\epsilon$ is a hyperparameter to clip the value function. Unlike other policy gradient methods, after collecting a new batch of rollout data, PPO continues to optimize the policy multiple times using Importance Sampling. 
The Importance Sampling ratio, $\frac{\pi\left(a_{t} | s_{t}\right)}{\pi_{\textrm{old}}\left(a_{t} | s_{t}\right)}$, can be interpreted as the probability of taking the given action under the current policy $\pi$ compared to the probability of taking the same action under the old policy that was used to generate the rollout data. 
The loss encourages the policy to take actions which have a positive advantage (better than average) while clipping discourages changes to the policy that overcompensate. This will effectively prevent the behavior of the current policy $\pi_{\theta}$ from deviating too far from the previous policy $\pi_{\theta_{old}}$.
PPO trains a value function $V(s_t)$ simultaneously with the policy in a supervised manner to estimate $\hat{V}_{t}$. In order to get the target value $\hat{V}_{t}$, we use Generalized Advantage Estimation (GAE)~\citep{Schulmanetal_ICLR2016}(Schulman et al, 2016).

\subsection{Soft Actor Critic (SAC)}
\label{SAC_method}
Since PPO is an on-policy algorithm, it inherently suffers from a data inefficiency problem. Soft Actor Critic (SAC) is a family of  data efficient off-policy algorithms which will allow for reuse of the already collected data from different scenarios. SAC is based on maximum entropy reinforcement learning, a method to simultaneously maximize both the expected reward and the entropy of the policy. Standard reinforcement algorithm aims to learn the parameters $\phi$ of some policy $\pi_{\phi} \left(a_{t} | s_{t}\right)$ such that the expected sum of rewards is maximized under the trajectory distribution $\rho_{\pi}$. However, the objective function of the maximum entropy algorithm also includes the entropy term $\mathcal{H}\left(\pi_{\phi}\left(\cdot | \mathbf{s}_{t}\right)\right)$ as a target to maximize. In short, it will converge to the policy that is the most random but still achieves a high reward. The objective function of maximum entropy reinforcement learning framework is shown as follows.

\begin{equation} \label{eq:2}
\phi^{*}=\arg \max _{\phi} \sum_{t=1}^{T} \underset{\left(\mathbf{s}_{t}, \mathbf{a}_{t}\right) \sim \rho_{\pi}}{\mathbb{E}}\left[r\left(\mathbf{s}_{t}, \mathbf{a}_{\iota}\right)+\alpha \mathcal{H}\left(\pi_{\phi}\left(\cdot | \mathbf{s}_{t}\right)\right)\right]
\end{equation}

where $r$ is the reward function, and $\alpha$ is a hyperparameter for tuning the trade-off between optimizing for the reward and for the stochasticy of the policy. In our experiments, $\alpha$ will be actively tuned with an automatic entropy tuning function~\citep{DBLP:journals/corr/abs-1812-05905}(Haarnoja et al, 2018). SAC is based on the soft policy iteration, a general algorithm for learning optimal maximum entropy policies that alternates between policy evaluation and policy improvement under the maximum entropy framework. Although original soft policy iteration only works on the discrete problems (i.e. tabular case), SAC extends the algorithm into continuous domain using function approximators for both the policy $\pi_{\phi}$ (actor) and the Q-function $Q_{\psi}$ (critic), and optimize these both network alternately. The soft Q-function parameters $\psi$ are optimized to minimize the soft Bellman residual,

\begin{equation} \label{eq:3}
J_{Q}(\psi)=\underset{\left(\mathbf{s}_{t}, \mathbf{a}_{t}, r_{t}, \mathbf{s}_{t+1}\right) \sim \mathcal{D}}{\mathbb{E}}\left[\frac{1}{2}\left(Q_{\psi}\left(\mathbf{s}_{t}, \mathbf{a}_{t}\right)-\left(r_{t}+\gamma V_{\overline{\psi}}\left(\mathbf{s}_{t+1}\right)\right)\right)\right]
\end{equation}

\begin{equation} \label{eq:4}
V_{\overline{\psi}}\left(\mathbf{s}_{t+1}\right)=\underset{\mathbf{a}_{t+1} \sim \pi_{\phi}}{\mathbb{E}}\left[Q_{\overline{\psi}}\left(\mathbf{s}_{t+1}, \mathbf{a}_{t+1}\right)-\alpha \log \pi_{\phi}\left(\mathbf{a}_{t+1} | \mathbf{s}_{t+1}\right)\right]
\end{equation}

where, $\mathcal{D}$ is the rollout memory buffer, $\gamma$ is the discount factor, and $\overline{\psi}$ is an exponentially moving average of the value network weights, which is to add stability to the training. Finally, the policy is updated targeting the exponential of the soft Q-function,

\begin{equation} \label{eq:5}
J_{\pi}(\phi)=\underset{\mathbf{s}_{t} \sim \mathcal{D}}{\mathbb{E}}\left[\underset{\mathbf{a}_{t} \sim \pi_{\phi}}{\mathbb{E}}\left[\alpha \log \left(\pi_{\phi}\left(\mathbf{a}_{t} | \mathbf{s}_{t}\right)\right)-Q_{\psi}\left(\mathbf{s}_{t}, \mathbf{a}_{t}\right)\right]\right]
\end{equation}

Benefit of the sample efficiency of the off-policy learning algorithm come from the fact that both value estimators and the policy can be trained entirely on off-policy data. Since SAC shows the better robustness and stability than other off-policy algorithm \citep{DBLP:journals/corr/abs-1801-01290}, we choose SAC as a second baseline algorithm for DoorGym.

\subsection{Realistic Rendering and Post Process using Unity}
\begin{figure}[t!]
    \centering
    \begin{subfigure}[b]{0.40\textwidth}
        \includegraphics[width=\textwidth]{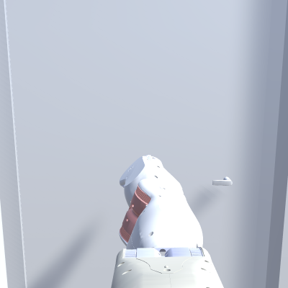}
        \caption{\scriptsize Original image}
        \label{segmentation_fig:before}
    \end{subfigure}
    \begin{subfigure}[b]{0.40\textwidth}
        \includegraphics[width=\textwidth]{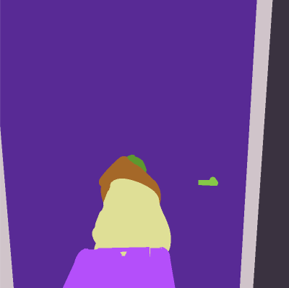}
        \caption{\scriptsize After the segmentation process}
        \label{segmentation_fig:after}
    \end{subfigure}
    \begin{subfigure}[b]{0.40\textwidth}
        \includegraphics[width=\textwidth]{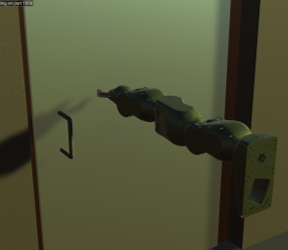}
        \caption{\scriptsize Original image}
        \label{segmentation_fig:original}
    \end{subfigure}
    \begin{subfigure}[b]{0.40\textwidth}
        \includegraphics[width=\textwidth]{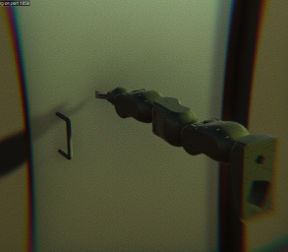}
        \caption{\scriptsize After adding the noises}
        \label{segmentation_fig:afternoise}
    \end{subfigure}
    \caption{Examples of the post processing that can be apply by Unity. Top row shows the function to make semantic segmentation labels. Bottom row shows the image when the Gaussian noise, camera distortion, and chromatic aberration are added.}
    \label{segmentation_fig}
\end{figure}

In order to improve transfer-ability and to increase domain randomization, the vision network can be trained with images that are closer to the real world image distribution. However, MuJoCo's built-in renderer is not designed to render photo-realistic images. OpenAI, 2018 \citep{openai2018learning} deal with this problem by using the Unity game engine as a rendering system. Official plugin for Unity parses MuJoCo world files and provides meshes, textures, etc., to Unity, and the simulation state is then synchronized between MuJoCo and Unity through TCP protocol. Our plugin extends the existing MuJoCo-Unity plugin to add additional control over the simulation, with the most noteworthy additions being support for loading worlds at run-time, randomization of material parameters, automatic UV-mapping of detail textures, and the output of semantic segmentation images. Image quality can be further improved with in-engine post-processing effects, such as ambient occlusion shaders which can approximate global lighting, lens distortion, depth of field, and image sensor noise. Post processing samples can be seen Figure \ref{segmentation_fig}.

\subsection{Vision Network Results}
\label{Vision Network Results}

To evaluate the necessity of both Unity's realistic rendering, and domain randomization for transfer, we perform two sets of experiments. First we evaluate the performance of MuJoCo/Unity models trained with and without domain randomization, both in the current domain, and in unseen domains. Second we compare models trained with data from MuJoCo with data from Unity to assess the transferability of each model.

As can be seen in table \ref{vision_net_results_tab} training the vision net with no domain randomization and trying to transfer to new domains gives vastly worse performance. MuJoCo gives an error of 0.29 cm in the same environment but explodes to anywhere between 19.89 and 35.82 cm when transferred to a new environment. This is contrasted by the model trained with DR which gives low out of error when tested in unseen MuJoCo environments. Transferring to completely different environments does cause error to go up substantially though, suggesting that DR is not enough alone for generalization. Unity shows a similar pattern of 1.71 cm in domain, with errors between 16.99 and 21.65 cm out of domain. DR once again helps, raising in domain error, but substantially reducing out of domain error.

\begin{figure}
    
    \centering
    \includegraphics[width=\textwidth]{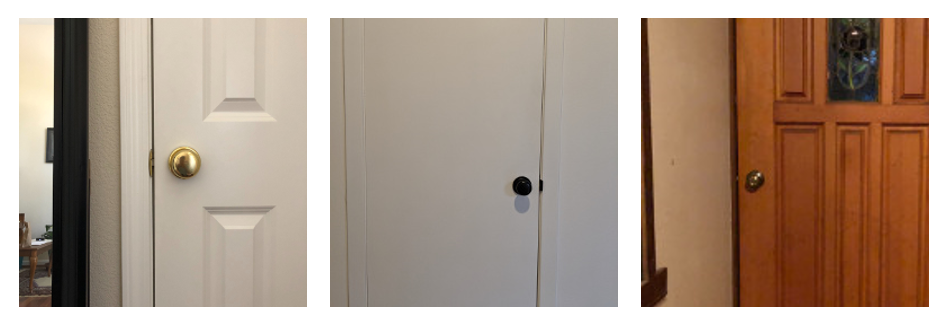}
    \caption{Sample knobs that used for sim2real transfer for the vision network.}
    \label{realdoor}
\end{figure}

For our second experiment with the vision network, we compare MuJoCO with DR to Unity with DR. As can be seen in table \ref{vision_net_results_tab}, each model has lower in domain error than when transferred, 0.43 cm as opposed to 3.49 cm when testing in unseen MuJoCo environments, and 10.33 cm compared to 3.41 cm when tested in Unity environments respectively. The gap between transferring MuJoCo to Unity is much larger (-9.90 cm) than Unity to MuJoCo (-0.08 cm). To fairly see the difference between using MuJoCo and Unity, we also transfer our results to real doorknobs, and see significantly lower error at 4.58 cm. We empirically noticed in our experiments that an error of 3 cm makes it difficult to open a door, but more than 5 cm made it impossible for the agent to locate the doorknob. Photos of the doorknobs that were used for real world transfer can be seen in Figure \ref{realdoor}.

\begin{table}[t]
\caption{Results of training and then evaluating vision network on MuJoCo and Unity rendered images with and without Domain Randomization (DR). All measurements are in cm. Bolded results are lowest error rate in that domain. Italicized are unseen environments for in the same domain as training.}
    \centering
    \begin{tabular}{@{}llllll@{}}
\toprule
               &       & \textbf{In Domain} & \textbf{Transfer to MuJoCo} & \textbf{Transfer to Unity} & \textbf{Transfer to Real} \\ \midrule
MuJoCo & No DR & \textbf{0.29}      & 19.89                       & 35.82                      & 23.47                     \\
                        & DR    & 0.43               & \textit{\textbf{0.43}}               & 10.33                      & 8.49                      \\ \midrule
Unity  & No DR & 1.71               & 16.99                       & 21.65                      & 19.53                     \\
                        & DR    & 3.41               & 3.49                        & \textit{\textbf{3.41}}              & \textbf{4.58}            
\end{tabular}
    \label{vision_net_results_tab}
\end{table}


\begin{table}[]
\caption{List of Randomized Parameters in the Door World}
    \label{tab:randomized_doorworld_parameters}
    \centering
    \begin{tabular}{@{}llll@{}}
    \toprule
    \multicolumn{2}{c}{\textbf{Randomization Parameters}} & \multicolumn{1}{c}{\textbf{Scalling Factor Range}} & \multicolumn{1}{c}{\textbf{Unit}} \\ \midrule
    \multicolumn{4}{c}{\textbf{Door Physical parameters}} \\ \midrule
    Wall Property & wall location & y-axis: Uniform[-200, 200] & mm \\
    Door Frame Joint Property & door frame damper & Uniform[0.1, 0.2] & - \\
     & door frame spring & Uniform[0.1,0.2] & - \\
     & door frame frictionloss & Uniform[0, 1] & - \\
    Door Property & door height & Uniform[2000, 2500] & mm \\
     & door width & Uniform[800, 1200] & mm \\
     & door thickness & Uniform[20, 30] & mm \\
     & knob height & Uniform[950, 1050] & mm \\
     & knob horizontal &  &  \\
     & location ratio & Uniform[0.10, 0.20] & - \\
     & door mass & Uniform[22.4, 76.5] & kg (Based on MDF density) \\
     & hinge position & left hinge/right hinge & - \\
     & opening direction & push/pull & - \\
    Knob Door Joint Property & knob door damper & Uniform[0.1, 0.2] & - \\
     & knob door spring & Uniform[0.1, 0.15] & - \\
     & knob door frictionloss & Uniform[0, 1] & - \\
     & knob rot range & Uniform[75, 80] & degree \\
    Knob Property & knob mass & Uniform[4, 7] & fg \\
     & knob surface friction & Uniform[0.50, 1.00] & - \\
     &  &  &  \\
    \multicolumn{4}{c}{\textbf{Robot Physical Parameters}} \\ \midrule
    Robot Property & arm joint damping & Uniform[0.1, 0.3] & - \\
     &  &  &  \\
    \multicolumn{4}{c}{\textbf{Vision Parameters}} \\ \midrule
    Lighting & light number & Uniform[2-6] & - \\
     & light diffuse & Uniform[0.0, 1.0] & RGBA \\
     & light position & x:Uniform[0.0, 5] & m \\
     &  & y:Uniform[-5, 5] &  \\
     &  & z:Uniform[3, 7] &  \\
     & light direction & x:Uniform[-0.5, 0.5] & rad \\
     &  & y:Uniform[-0.5, 5.0] &  \\
     &  & z:Uniform[-0.5, -0.25] &  \\
    Wall Material & wall shininess & Uniform[0.01, 0.50] & - \\
     & wall specular & Uniform[0.01, 0.50] & - \\
     & wall rgb1 & Uniform[0.0, 1.0] & RGBA \\
    Frame Material & frame shininess & Uniform[0.01, 0.70] & - \\
     & frame specular & Uniform[0.01, 0.80] & - \\
     & frame rgb1 & Uniform[0.0, 1.0] & RGBA \\
    Door Material & door shininess & Uniform[0.01, 0.30] & - \\
     & door specular & Uniform[0.01, 0.80] & - \\
     & door rgb1 & Uniform[0.0, 1.0] & RGBA \\
     & door rgb2 & Uniform[0.0, 1.0] & RGBA \\
    Doorknob Material & knob shininess & Uniform[0.50, 1.00] & - \\
     & knob specular & Uniform[0.80, 1.00] & - \\
     & knob rgb1 & Uniform[0.0, 1.0] & RGBA \\
    Robot Material & robot shininess & Uniform[0.01, 0.70] & - \\
     & robot specular & Uniform[0.01, 0.80] & - \\
     & robot rgb & Uniform[0.0, 1.0] & RGBA \\ \bottomrule
    \end{tabular}
\end{table}

\begin{table}[]
\caption{Hyperparameters for PPO}
    \label{tab:PPO_hyperparameters}
    \centering
    \begin{tabular}{@{}ll@{}}
    \toprule
    \multicolumn{1}{c}{\textbf{Hyperparameter}} & \textbf{Value}                   \\ \midrule
   {Hardware Configuration}     & 8 NVIDIA GeForce GTX TITAN-X GPU \\
                                                & 8 CPU cores                      \\ \midrule
    Optimizer                                   & Adam \citep{Adam}(Kingma et al, 2014)                \\ \midrule
    Learning Rate                               & 0.001 \\ \midrule
    Multi-worker number                         & 8                                \\ \midrule
    PPO mini-batch size                         & 256                              \\ \midrule
    PPO clipping ratio                          & 0.2                              \\ \midrule
    Max Grad Norm                               & 0.5                              \\ \midrule
    Discount Ratio Gamma                        & 0.99                             \\ \midrule
    GAE lambda                                  & 0.95                             \\ \midrule
    Entropy-coef                                & 0.0                              \\ \midrule
    Action-loss-coef                            & 1.0                              \\ \midrule
    Value-loss-coef                             & 0.5                              \\ \bottomrule
    \end{tabular}
\end{table}

\begin{table}[]
\caption{Hyperparameters for SAC}
    \label{tab:SAC_hyperparameters}
    \centering
    \begin{tabular}{@{}ll@{}}
    \toprule
    \multicolumn{1}{c}{\textbf{Hyperparameter}} & \textbf{Value}                   \\ \midrule
   {Hardware Configuration}     & NVIDIA GeForce GTX TITAN-X GPU \\
                                                & 8 CPU cores                      \\ \midrule
    Optimizer                                   & Adam \citep{Adam}                \\ \midrule
    Policy Learning Rate                        & 0.001                            \\ \midrule
    Q-function Learning Rate                    & 0.001                            \\ \midrule
    Discount Ratio Gamma                        & 0.99                             \\ \midrule
    Rollout Memory Buffer Size                  & $10^6$                             \\ \midrule
    Target update Period                        & 1                                \\ \midrule
    Target smoothing coef                       & 0.005                            \\ \midrule
    Automatic Entorpy Tuning                    & True                              \\ \midrule
    \end{tabular}
\end{table}

\begin{figure}[t]
    \centering
    \includegraphics[width=\textwidth]{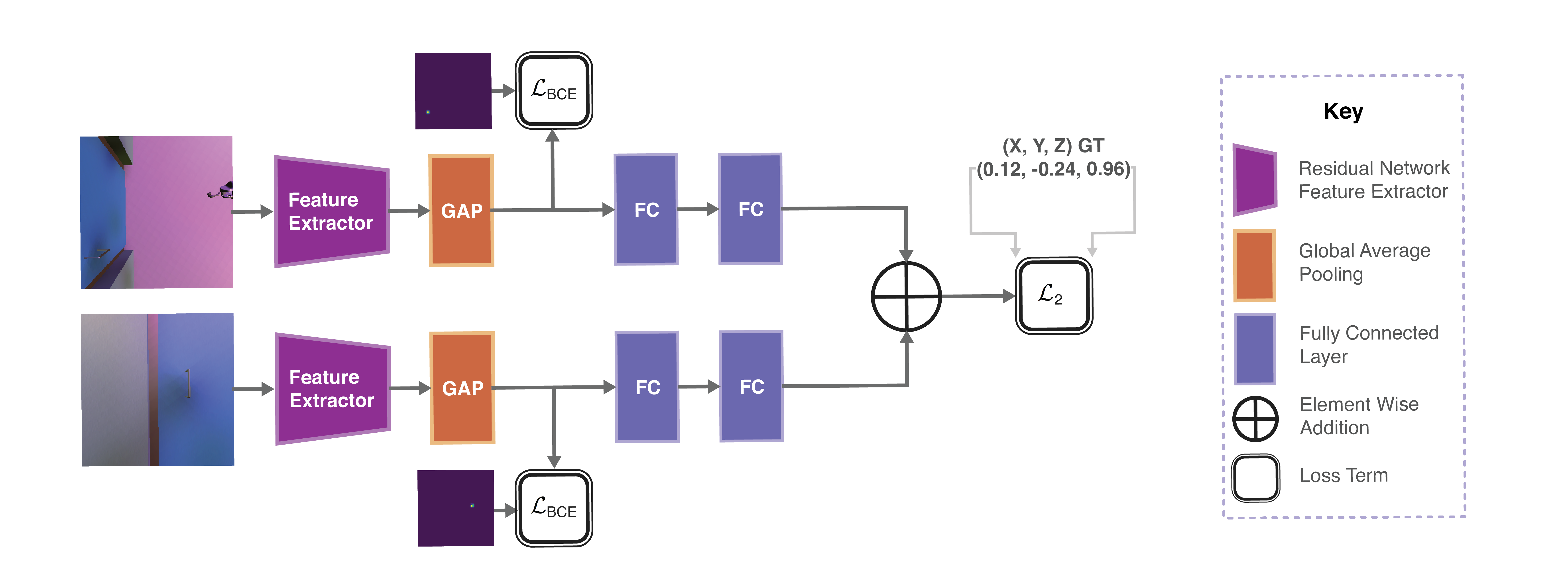}
    \caption{Vision Network Architecture}
    \label{fig:vision_network}
\end{figure}

\begin{figure}[t]
    \centering
    \includegraphics[width=0.8\textwidth]{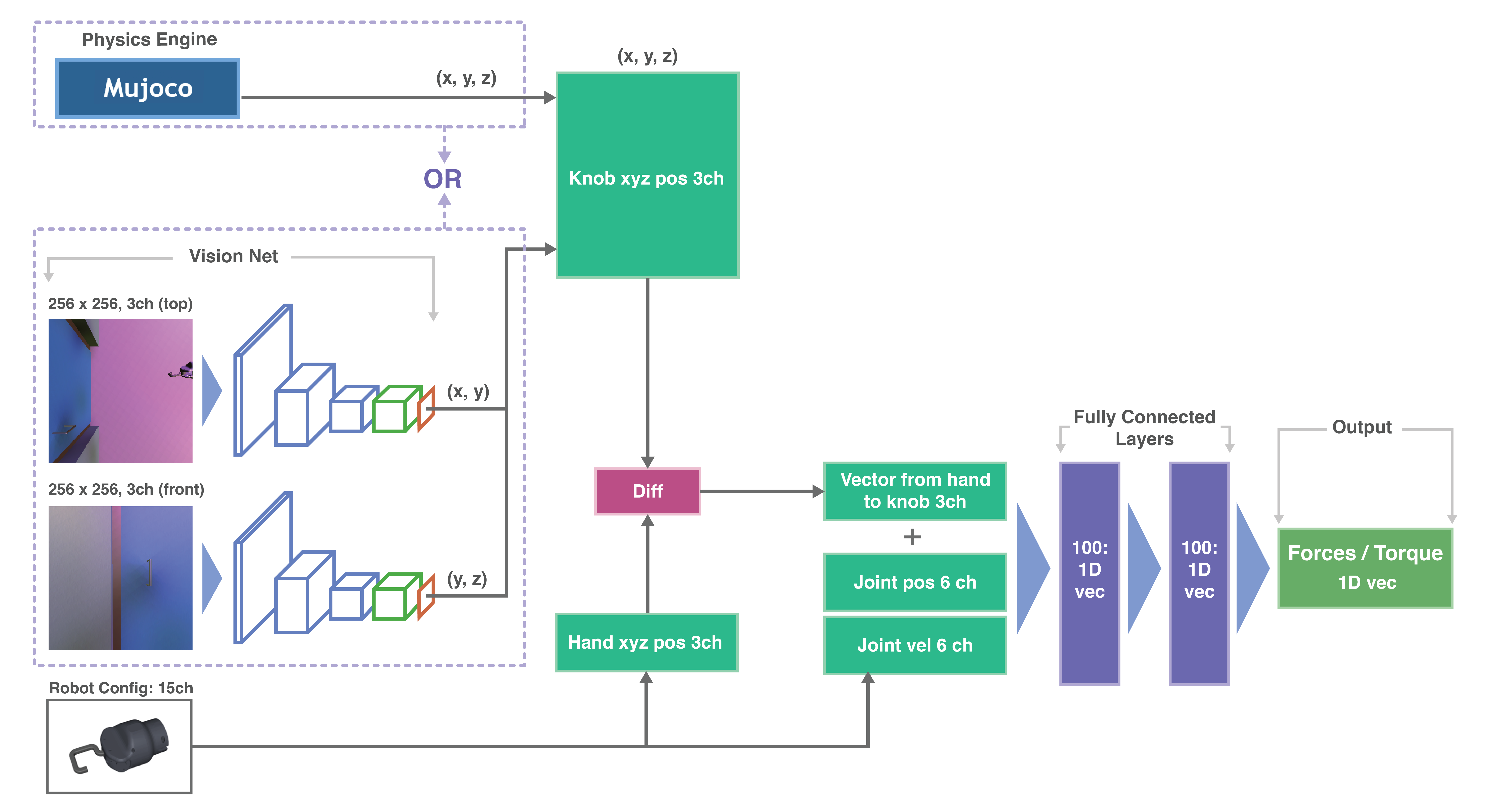}
    \caption{Policy Network Architecture}
    \label{fig:policyarch}
\end{figure}

\begin{table}[]
\caption{Result of experiments using PPO}
    \label{tab:full_result_PPO}
    \centering
\begin{tabular}{@{}cccccccccc@{}}
\toprule
\multicolumn{1}{l}{\textbf{Algorithm}} & \multicolumn{1}{l}{\textbf{Open Direction}} & \multicolumn{1}{l}{\textbf{Robot type}} & \multicolumn{1}{l}{\textbf{Knob type}} & \multicolumn{2}{c}{\textbf{GT}} & \multicolumn{2}{c}{\textbf{GT + Noise}} & \multicolumn{2}{c}{\textbf{Vision}} \\ \cmidrule(l){5-10} 
\textbf{} & \textbf{} & \textbf{} & \textbf{} & $r_{\textrm{ASR}}$ & $r_{\textrm{AT}}$ & $r_{\textrm{ASR}}$ & $r_{\textrm{AT}}$ & $r_{\textrm{ASR}}$ & $r_{\textrm{AT}}$ \\ \midrule
PPO & pull & hook & pull & 0.95 & 3.47 & 0.99 & 3.20 & 0.79 & 3.73 \\
 &  &  & lever & 0 & - & 0 & - & 0 & - \\
 &  &  & round & 0.51 & 5.56 & 0.47 & 5.20 & 0.26 & 5.47 \\ \cmidrule(l){3-10} 
 &  & gripper & pull & 0.68 & 4.18 & 0.59 & 4.12 & 0.62 & 3.73 \\
 &  &  & lever & 0 & - & 0 & - & 0 & - \\
 &  &  & round & 0 & - & 0 & - & 0 & - \\ \cmidrule(l){3-10} 
 &  & floating hook & pull & 0.95 & 4.51 & 0.78 & 5.18 & 0.48 & 6.15 \\
 &  &  & lever & 0.68 & 4.12 & 0.54 & 4.11 & 0.29 & 3.83 \\
 &  &  & round & 0 & - & 0 & - & 0 & - \\ \cmidrule(l){3-10} 
 &  & floating gripper & pull & 0.93 & 4.81 & 0.86 & 5.49 & 0.55 & 5.79 \\
 &  &  & lever & 0 & - & 0 & - & 0 & - \\
 &  &  & round & 0 & - & 0 & - & 0 & - \\ \cmidrule(l){3-10} 
 &  & mobile hook & pull & 0.84 & 4.41 & 0.8 & 4.88 & 0.66 & 5.36 \\
 &  &  & lever & 0.88 & 5.82 & 0.67 & 5.46 & 0.37 & 5.51 \\
 &  &  & round & 0 & - & 0 & - & 0 & - \\ \cmidrule(l){3-10} 
 &  & mobile gripper & pull & 0.71 & 5.13 & 0.65 & 4.73 & 0.57 & 5.12 \\
 &  &  & lever & 0 & - & 0 & - & 0 & - \\
 &  &  & round & 0 & - & 0 & - & 0 & - \\ \cmidrule(l){2-10} 
 & push & hook & pull & 1.00 & 2.01 & 1.00 & 1.53 & 1.00 & 1.52 \\
 &  &  & lever & 0.11 & 6.94 & 0.04 & 4.92 & 0 & - \\
 &  &  & round & 0.11 & 7.45 & 0.14 & 6.65 & 0 & - \\ \cmidrule(l){3-10} 
 &  & gripper & pull & 1.00 & 1.20 & 1.00 & 1.01 & 1.00 & 1.11 \\
 &  &  & lever & 0.10 & 5.04 & 0.21 & 5.71 & 0.21 & 4.29 \\
 &  &  & round & 0.04 & 7.43 & 0 & - & 0 & - \\ \cmidrule(l){3-10} 
 &  & floating hook & pull & 0.97 & 2.02 & 1.00 & 2.22 & 1.00 & 2.14 \\
 &  &  & lever & 0.37 & 5.38 & 0 & - & 0 & - \\
 &  &  & round & 0.02 & 2.63 & 0.02 & 4.75 & 0.02 & 4.14 \\ \cmidrule(l){3-10} 
 &  & floating gripper & pull & 0.98 & 2.37 & 1.00 & 2.19 & 1.00 & 2.31 \\
 &  &  & lever & 0.66 & 6.99 & 0.16 & 7.12 & 0.05 & 8.15 \\
 &  &  & round & 0 & - & 0.01 & 2.2 & 0 & - \\ \cmidrule(l){3-10} 
 &  & mobile hook & pull & 1.00 & 2.71 & 1.00 & 1.84 & 1.00 & 1.64 \\
 &  &  & lever & 0.44 & 5.22 & 0.45 & 4.62 & 0.39 & 4.11 \\
 &  &  & round & 0.06 & 7.70 & 0.12 & 6.29 & 0.06 & 5.73 \\ \cmidrule(l){3-10} 
 &  & mobile gripper & pull & 0.98 & 1.68 & 1.00 & 1.38 & 1.00 & 1.42 \\
 &  &  & lever & 0.56 & 5.67 & 0.63 & 5.15 & 0.41 & 5.23 \\
 &  &  & round & 0.25 & 8.47 & 0.22 & 6.96 & 0.08 & 8.66 \\ \bottomrule
\end{tabular}
\end{table}

\begin{table}[]
\caption{Result of experiments using SAC}
    \label{tab:full_result_SAC}
    \centering
\begin{tabular}{@{}cccccccc@{}}
\toprule
\multicolumn{1}{l}{\textbf{Algorithm}} & \multicolumn{1}{l}{\textbf{Open Direction}} & \multicolumn{1}{l}{\textbf{Robot type}} & \multicolumn{1}{l}{\textbf{Knob type}} & \multicolumn{2}{c}{\textbf{GT}} & \multicolumn{2}{c}{\textbf{Vision}} \\ \cmidrule(l){5-8} 
\textbf{} & \textbf{} & \textbf{} & \textbf{} & $r_{\textrm{ASR}}$ & $r_{\textrm{AT}}$ & $r_{\textrm{ASR}}$ & $r_{\textrm{AT}}$ \\ \midrule
SAC & pull & hook & pull & 0.62 & 3.21 & 0.25 & 2.93 \\
 &  &  & lever & 0 & - & 0 & - \\
 &  &  & round & 0 & - & 0 & - \\ \cmidrule(l){3-8} 
 &  & gripper & pull & 0.58 & 3.42 & 0.07 & 2.36 \\
 &  &  & lever & 0 & - & 0 & - \\
 &  &  & round & 0.01 & 7.08 & 0 & - \\ \cmidrule(l){3-8} 
 &  & floating hook & pull & 0.23 & 4.29& 0 & - \\
 &  &  & lever & 0 & - & 0 & - \\
 &  &  & round & 0.03 & 7.21 & 0 & - \\ \cmidrule(l){3-8} 
 &  & floating gripper & pull & 0.47 & 4.98 & 0 & - \\
 &  &  & lever & 0.19 & 7.99 &  0 & - \\
 &  &  & round & 0 & - & 0 & - \\ \cmidrule(l){3-8} 
 &  & mobile hook & pull & 0.33 & 5.02 & 0.12 & 3.72 \\
 &  &  & lever & 0.05 & 5.93 & 0 & - \\
 &  &  & round & 0.03 & 5.56 & 0 & - \\ \cmidrule(l){3-8} 
 &  & mobile gripper & pull & 0.47 & 4.03 & 0.05 & 5.06 \\
 &  &  & lever & 0.13 & 6.39 & 0.01 & 6.94 \\
 &  &  & round & 0 & - & 0 & - \\ \cmidrule(l){2-8} 
 & push & hook & pull & 1.00 & 1.24 & 0.99 & 0.76 \\
 &  &  & lever & 0.09 & 5.71 & 0.01 & 3.12 \\
 &  &  & round & 0.04 & 3.56 & 0.01 & 9.66 \\ \cmidrule(l){3-8} 
 &  & gripper & pull & 0.99 & 1.04 & 0.99 & 0.80 \\
 &  &  & lever & 0 & - & 0 & - \\
 &  &  & round & 0.01 & 7.46 & 0 & - \\ \cmidrule(l){3-8} 
 &  & floating hook & pull & 0.99 & 2.22 & 0.99 & 1.95 \\
 &  &  & lever & 0.02 & 8.62 & 0 & - \\
 &  &  & round & 0.02 & 7.2 & 0 & - \\ \cmidrule(l){3-8} 
 &  & floating gripper & pull & 0.98 & 2.04 & 0.98 & 2.15 \\
 &  &  & lever & 0.43 & 6.68 & 0 & - \\
 &  &  & round & 0 & - & 0 & - \\ \cmidrule(l){3-8} 
 &  & mobile hook & pull & 1.00 & 1.26 & 0.95 & 1.23 \\
 &  &  & lever & 0.23 & 6.03 & 0.05 & 6.89 \\
 &  &  & round & 0.03 & 2.92 & 0.03 & 2.23 \\ \cmidrule(l){3-8} 
 &  & mobile gripper & pull & 0.97 & 1.12 & 0.98 & 1.03 \\
 &  &  & lever & 0.21 & 6.74 & 0.12 & 5.07 \\
 &  &  & round & 0.01 & 9.98 & 0.02 & 9.25 \\ \bottomrule
\end{tabular}
\end{table}

\end{document}